\documentclass{article}
\pdfoutput=1

\usepackage{PRIMEarxiv}

\usepackage[utf8]{inputenc} 
\usepackage[T1]{fontenc}    
\usepackage{hyperref}       
\usepackage{url}            
\usepackage{booktabs}       
\usepackage{amsfonts}       
\usepackage{nicefrac}       
\usepackage{microtype}      
\usepackage{lipsum}
\usepackage{fancyhdr}       
\usepackage{graphicx}       
\graphicspath{{media/}}     

\usepackage{amssymb}
\usepackage{amsmath}
\usepackage{lineno}
\usepackage{float}
\usepackage{array}
\usepackage{subfigure}
\usepackage{bm}
\usepackage{tikz}
\usepackage{multirow}
\usepackage{threeparttable}
\usepackage{setspace}

\title{A data-based comparative review \\ and AI-driven symbolic model \\ for longitudinal dispersion coefficient in natural streams   

}

\author{ \href{https://orcid.org/0000-0002-2247-4376}{\includegraphics[scale=0.06]{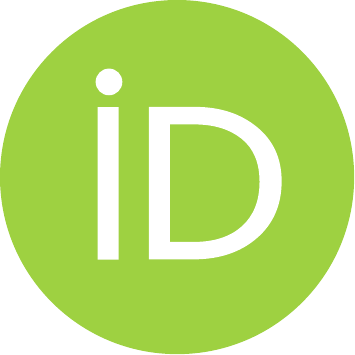}\hspace{1mm}Yifeng Zhao} \\
	Department of Environmental Science and Engineering\\
	Zhejiang University\\
	Hangzhou, CHN \\
    \& \\
    School of Engineering \\
	Westlake University\\
    Hangzhou, CHN\\
	\texttt{zhaoyifeng@westlake.edu.cn} \\
	\And
	\href{}{\includegraphics[scale=0.06]{orcid.pdf}\hspace{1mm}Zicheng Liu} \\
	School of Engineering \\
	Westlake University\\
    Hangzhou, CHN\\
	\texttt{liuzicheng@westlake.edu.cn} \\
    \AND
    \href{https://orcid.org/0000-0002-0965-1374}{\includegraphics[scale=0.06]{orcid.pdf}\hspace{1mm}Pei Zhang} \\
	School of Engineering \\
	Westlake University\\
    Hangzhou, CHN\\
	\texttt{zhangpei@westlake.edu.cn} \\
    \And
    \href{}{\includegraphics[scale=0.06]{orcid.pdf}\hspace{1mm}S.A. Galindo-Torres}\thanks{Co-corresponding author}\\
	School of Engineering \\
	Westlake University\\
    Hangzhou, CHN\\
	\texttt{s.torres@westlake.edu.cn} \\
    \And
    \href{https://orcid.org/0000-0002-2961-8096}{\includegraphics[scale=0.06]{orcid.pdf}\hspace{1mm}Stan Z. Li}\thanks{Co-corresponding author}\\
	School of Engineering \\
	Westlake University\\
    Hangzhou, CHN\\
	\texttt{stan.zq.li@westlake.edu.cn} \\

}

\begin{document}
\maketitle

\begin{abstract}
  A better understanding of contaminant spills in natural streams requires knowledge of Longitudinal Dispersion Coefficient (LDC). 
  Various methods have been proposed for predictions of LDC. Those studies can be grouped into three types: 
  analytical, statistical, and Machine Learning (ML) driven researches. 
  However, a comprehensive evaluation of them is still lacking. 
  In this paper, 
  we first present an in-depth analysis of those methods and reveal some of their drawbacks. 
  This is carried out on an extensive database composed of 660 samples 
  of hydraulic and channel properties worldwide. The reliability and representativeness of the data 
  are enhanced by deploying the Subset Selection of Maximum Dissimilarity (SSMD) for 
  separation of subsets and the 
  Inter Quartile Range (IQR) for removal of outliers. 
  The analysis reveals the rank of those different methods as: ML-driven method $>$ the statistical method $>$ 
  the analytical method. Where the '$>$' represents greater predictive performance.  
  To establish an interpretable model for LDC prediction with higher performance, 
  we design a novel interpretable 
  regression method called evolutionary symbolic regression 
  network (ESRN). It is a combination of genetic algorithms and neural networks. 
  Strategies are introduced to avoid overfitting and explore more parameter combinations. 
  Results show that the ESRN model distilled with a larger dataset and better processing strategies
  has superiorities over other existing symbolic models in performance and reliability. 
  Also, a striking finding is that ESRN produced a simpler formula with a smaller number of 
  parameters than the existing alternatives. 
  Its simplicity allows us to connect this relation to the fundamental turbulent mixing process, 
  illustrating this is the underlying physics behind the LDC in rivers.
  The proposed model is suitable for practical engineering problems and can provide convincing 
  solutions for situations where the field test cannot be carried out, or limited field information can be obtained.

\end{abstract}


\section{Introduction}\label{sec:intro}

Natural streams play a pivotal role in the cycle of water resources on the earth. 
They have a strong impact on human water utilization processes, such as agriculture, 
ecology, cities, and industry. Meanwhile, they receive different kinds of contamination 
during these processes. Some of them, especially chemical and waste 
compounds, are crucial to the quality assessment of streams. In recent years, frequent contaminant 
spills have caused significant public attention. 
In Norilsk of Russia (2020), more than 135 square miles of the area have been 
contaminated due to an oil spill\cite{russianPollution}. In Fukushima of Japan (2019) 
the report of the nuclear waste leak has caused 
great panic around the pacific rim\cite{JapanNuclear}. 
In West Virginia of USA (2014), more than 300,000 people were affected 
by a toxic compound spill\cite{USAchemical}. Among these cases, the 
pollution will spread throughout the ecosystem, 
not only to the lower algae animals but also to the human level. Recent studies have 
shown that the stream contamination is a potential cause 
of various diseases to both humans and other mammals\cite{oguntoke2010association, malins1984, identifying2021}. 
Without an accurate tool 
to quantify this process, it is challenging to assess the quality of the stream and 
the overall pollution status, which makes it more 
time-consuming to fully solve this problem and restore the environment 
to its original state. On top of this, it is often challenging 
to determine the specific scope of impact with present techniques.

To control and reduce the damage of similar water pollution incidents, an 
effective method is to use simulation technology to evaluate 
the overall water status, which is faster and covers a broader range than the 
traditional sampling detection method. However, the existing 
simulation algorithms developed according to established theories have revealed 
mismatches through comparison of the sampling test 
method in several applications\cite{ramezani2019numerical}. To improve accuracy, 
it is vital to obtain a deeper understanding of this phenomenon. 

When scalars enter the stream, a 3-D process will be involved. In the early 
stage of the entrance, the advection will be 
dominant due to flow velocity for a short period of time. 
After the scalar is fully mixed, a balance between advection and diffusion 
will be reached. Then the longitudinal advection and 
dispersion start to reign. The general Advection 
Diffusion Equation (ADE, Eq. \ref{eq:ADE}) can be used for further description. 

\begin{equation}
    \frac{\partial{C}}{\partial{t}}= \nabla \cdot (D_l \nabla C) - \nabla \cdot (U C) + R
    \label{eq:ADE}
\end{equation}
Where C - the concentration; t - the time; $D_l$ - the longitudinal dispersion coefficient; U - the flow velocity; R - the reaction term
signaling either sources or sinks of the solute.

As for solutions to the ADE, both analytical and numerical methods have been developed 
in the past\cite{bloschl1995scale, fischer1979mixing, alosairi2020three}. These studies showed that $D_l$
is the most significant parameter for the accuracy of ADE solutions\cite{fischer1966longitudinal}. 
The field measurement of $D_l$ can be costly and time-consuming. Therefore, 
many attempts have been made on the prediction of $D_l$ with its wide range of influence 
variables. Major methods for prediction are the analytical, statistical, 
and the Machine Learning driven method. However, the present methods to obtain $D_l$ cannot fully satisfy 
the need for practical engineering\cite{malins1984,noori2011a,memarzadeh2020a}. 
Those methods suffer from several disadvantages, such as weak prediction, 
poor generalization, and \textit{black-box} nature (The property of algorithms and techniques that give no explicit 
explanation of working mechanisms). 

To remedy this, a novel evolutionary symbolic regression network is proposed 
to distill predictive equations for $D_l$ from extensive field data. 
The resulting $D_l$ formula shows superiority in 
accuracy and generalization than previous researches. 
It also indicates the underlying physics behind the LDC is 
the turbulent mixing process. 

The remainder of this paper is organized as follows: 
Section \ref{sec:previousStudies} is a detailed review of 
past research on the prediction of $D_l$. Section \ref{sec:data} consists of three parts: 
\emph{Data exploratory and pre-process}, \emph{Training and testing sets} and 
\emph{Evaluation on previous studies}. \emph{Data exploratory and pre-process} illustrates the strong data basis of 
this paper, a dataset of 1094 samples. Pre-processes, especially the 
removal of outliers with 
Inter Quartile Range (IQR), are also carried out in this part.
\emph{Training and testing sets} refers to the use 
of the Subset Selection of Maximum Dissimilarity (SSMD) to separate the dataset into the
training set and testing set with the same distribution. \emph{Evaluation on previous studies} presents 
a comparison between different models and methods for the prediction of $D_l$. 
Section \ref{sec:method} reveals the theory and design of the novel symbolic regression algorithm used in this study. 
In Section \ref{sec:result}, the novel $D_l$ prediction equation is presented. Multiple techniques are used to evaluate its
performance and the result reveals its superiority over other research. Finally, 
Section \ref{sec:conclusion} summarizes the contribution
and indicates possible improvements for future studies.

\section{Previous studies on the prediction of $D_l$}\label{sec:previousStudies}
$D_l$ has many potential impact factors, such as stream properties, hydraulic conditions, and 
channel geometrics. According to turbulence properties of flow and 
simplification in flow-path parameters, it is found that the most important variables are:
the average channel width - $w$, the average channel depth - $d$, the flow velocity - $U$, and 
the shear velocity - $U^*$\cite{IlWonSeo.1998, ZhiQiangDeng.2001,
alizadeh2017predicting, alizadeh2017improvement}. In this framework, many studies 
have been carried out. Studies on this topic can be divided into three categories: 
analytical, statistical, and the ML-driven method. 
 
The analytical method combines mathematical principles and conceptual abstractions to obtain an approximate prediction. 
It is the first solution applied to this 
problem and several successful cases have already been achieved under ideal conditions, 
such as laminar flow in a pipe\cite{taylor1953dispersion}, turbulent circular flow in 
a pipe\cite{taylor1954the}, and open channel flow\cite{elder1959the, fisher1968dispersion,fischer1966longitudinal}. 
Some of those 
studies are listed in Table \ref{T:analyticalEq}.

\begin{table}[]\centering
    \caption{Summary of analytical model on $D_l$}
    \resizebox{\textwidth}{!}{
    \begin{threeparttable}
    \begin{tabular}{cccc}
    \hline
    \multicolumn{1}{|c|}{Seq}   &\multicolumn{1}{|c|}{Author / Year}  & \multicolumn{1}{c|}{The formula}                                                                         & \multicolumn{1}{c|}{Application}                         \\ \hline
    \multicolumn{1}{|c|}{1}   &\multicolumn{1}{|c|}{Taylor / 1953}  & \multicolumn{1}{c|}{$D_l=\frac{(aU)^2}{48D_p}$}                                                            & \multicolumn{1}{c|}{The laminar pipe flow}               \\ \hline
    \multicolumn{1}{|c|}{2}   &\multicolumn{1}{|c|}{Taylor / 1953}  & \multicolumn{1}{c|}{$D_l=10.06au_*$}                                                                       & \multicolumn{1}{c|}{The turbulent circular pipe flow}    \\ \hline
    \multicolumn{1}{|c|}{3}   &\multicolumn{1}{|c|}{Elder / 1959}   & \multicolumn{1}{c|}{$D_l=(\frac{0.4041}{\kappa^3}+\frac{\kappa}{6})dU^* \quad or \quad D=5.86dU^*$}        & \multicolumn{1}{c|}{The open channel flow}               \\ \hline
    \multicolumn{1}{|c|}{4}   &\multicolumn{1}{|c|}{Fischer / 1968} & \multicolumn{1}{c|}{$D_l = \int_{0}^{w}du\prime \int_{0}^{y}\frac{1}{\epsilon_t d} \int_{0}^{y}du\prime \,dy\,dy\,dy$}                  & \multicolumn{1}{c|}{The open channel flow}               \\ \hline

    \end{tabular}%

    \begin{tablenotes}
        \item[*] $D_l$ - the longitudinal dispersion coefficient; $a$ - The radius of the pipe; $D_{p}$ - The diffusion coefficient of particles; $u_{*}$ - The shear velocity of circular pipe flow, 
                 equal to $(\frac{gaS}{2})^{0.5}$, where $S$ is the energy slope; 
                 $U$ - The flow velocity; $U^*$ - The shear velocity of natural streams, equal to $(gRS)^{0.5}$, where $R$ is the hydraulic radius; $w$ - the average channel width; $d$ - the average channel depth; $\kappa$ - the von Karman constant, approximately equal to 0.41;
                 $u\prime$ - the deviation of the velocity from the cross-sectional mean velocity; $y$ - Cartesian coordinate in the lateral direction; $\epsilon_t$ - the transverse turbulent diffusion coefficient. 
        
    \end{tablenotes}
    \end{threeparttable}}
    \label{T:analyticalEq}
\end{table}

Although those studies are still used in many engineering problems today, they suffer from several defects. 
A good example is that those prediction 
formulas are obtained under certain assumptions. The demand in detailed flow properties ($D_{p}$, $S$) 
and cross-sectional information ($y$, $\epsilon_t$) is also hard to satisfy. 
Besides, forms of some equations (Fischer/1978, Taylor / 1953 and Elder/1959) are overly complicated. 
These drawbacks restrict their application in more complex engineering problems.

To remedy those weakness, Fischer\cite{fischer1966longitudinal} 
applied statistical methods on the research of $D_l$. By introducing
four macro-variables: $d$, $w$, $U$ and $U^*$, the prediction equation is simplified and
the $D_l$ is connected to its geometric and hydrologic variables (Eq. \ref{eq:fischerSimplyf}).

\begin{equation}
\frac{D_l}{dU^*}=0.011 \left(\frac{w}{d} \right)^2 \left(\frac{U}{U^*} \right)^2
\label{eq:fischerSimplyf}
\end{equation}

After that, through use of various statistical methods and expansion of the $D_l$ dataset, 
a series of studies are carried out (Table \ref{T:statisticalEq}). Utilizing the similarity between the 
1D flow equation and 1D dispersion equation,
McQuivey and Keefer\cite{mcquivey1974simple} developed a simplified equation for $D_l$ problem with Froude number ($Fr$) less than 0.5, 
where the Froude Number is defined as: 
\begin{equation}
    \mathrm{Fr}=\frac{U}{\sqrt{g L}}
    \label{eq:froude}
\end{equation}
    where $g$ - the local external field; $L$ - the characteristic length. 

Liu et al.\cite{liu1977predicting} modified Fischer's equation (Eq. \ref{eq:fischerSimplyf}) to a form which contains a gradient of lateral velocity. Through fitting,
a highly simplified formula was proposed;
Dimensional analysis was applied to this problem by Seo and Cheong for the first time\cite{seo1998predicting}. The one-step Huber method,
a famous nonlinear multi-regression method was then utilized on 59 field datasets to enhance the previous research;
Koussis and Rodríguez-Mirasol applied Von Karmen defect law to
the original theory and equations formulated by Fischer(Eq. \ref{eq:fischerSimplyf}) and a modified equation was developed\cite{koussis1998hydraulic};
A theoretical approximation to generate a new equation based on Fischer work was made by Deng et al., and a more detailed expression of the channel
depth, the flow velocity, and the local transverse mixing coefficient was proposed\cite{deng2001longitudinal};
Through the expansion of dataset, Kashefipour and Falconer\cite{kashefipour2002longitudinal}, Zeng and Huai\cite{zeng2014estimation}
both presented improved $D_l$ prediction formulas.
Disley et al. incorporated the Froude number as a third key parameter other than w/d and U/U*.
This new combination achieved better results in the testing\cite{disley2015predictive}.
Those studies grasped the essence of the $D_l$ prediction, a regression problem. More simple and practical results have been
obtained with the use of different statistical regression methods and introduce of new variables.

\begin{table}[]
    \caption{Summary of statistical model on $D_l$}
    \resizebox{\textwidth}{!}{
    \begin{threeparttable}
    \begin{tabular}{cccc}
    \hline
    \multicolumn{1}{|c|}{Seq}   &\multicolumn{1}{|c|}{Author / Year}                                          & \multicolumn{1}{c|}{The formula}                                                                                                                                                                             & \multicolumn{1}{c|}{Application}                         \\ [3pt]\hline 
    \multicolumn{1}{|c|}{5}   &\multicolumn{1}{|c|}{Fischer / 1979}                                         & \multicolumn{1}{c|}{$\frac{D_l}{dU^*} = 0.011(\frac{w}{d})^2(\frac{U}{U^*})^2$}                                                                                                                              & \multicolumn{1}{c|}{\begin{tabular}[c]{@{}c@{}}Connect $D_l$ with its macro-variables with\\ statistical methods for the first time\end{tabular}}               \\ [3pt]\hline
    \multicolumn{1}{|c|}{6}   &\multicolumn{1}{|c|}{McQuivey and Keefer / 1975}                             & \multicolumn{1}{c|}{$\frac{D_l}{dU^*} = 0.058(\frac{U}{SU^*})$}                                                                                                                                              & \multicolumn{1}{c|}{For Froude number less than 0.5}    \\ [3pt]\hline
    \multicolumn{1}{|c|}{7}   &\multicolumn{1}{|c|}{Liu et al. / 1977}                                      & \multicolumn{1}{c|}{$\frac{D_l}{dU^*} = 0.18(\frac{w}{d})^2(\frac{U}{U^*})^{0.5}$}                                                                                                                           & \multicolumn{1}{c|}{Consider lateral velocity gradient}               \\ [3pt]\hline
    \multicolumn{1}{|c|}{8}   &\multicolumn{1}{|c|}{Seo and Cheong / 1998}                                  & \multicolumn{1}{c|}{$\frac{D_l}{dU^*} = 5.915(\frac{w}{d})^{0.62}(\frac{U}{U^*})^{1.428}$}                                                                                                                   & \multicolumn{1}{c|}{Apply von Karmen defect law}               \\ [3pt]\hline
    \multicolumn{1}{|c|}{9}   &\multicolumn{1}{|c|}{Koussis and Rodríguez-Mirasol / 1998}                   & \multicolumn{1}{c|}{$\frac{D_l}{dU^*} = 0.6(\frac{w}{d})^2$}                                                                                                                                                 & \multicolumn{1}{c|}{Apply dimensional analysis}               \\ [3pt]\hline
    \multicolumn{1}{|c|}{10}   &\multicolumn{1}{|c|}{Deng et al. / 2001}                                     & \multicolumn{1}{c|}{$\frac{D_l}{dU^*} = (\frac{0.15}{8\epsilon})(\frac{w}{d})^{1.67}(\frac{U}{U^*})^2$}                                                                                                      & \multicolumn{1}{c|}{\begin{tabular}[c]{@{}c@{}}Introduce a new variable with\\ detailed theoretical analysis\end{tabular}}               \\ [3pt]\hline
    \multicolumn{1}{|c|}{\multirow{2}{*}{11}}   &\multicolumn{1}{|c|}{\multirow{2}{*}{Kashefipour and Falconer / 2002}}       & \multicolumn{1}{c|}{\multirow{2}{*}{\begin{tabular}[c]{@{}c@{}}$\frac{w}{d}>50,$ $\frac{D_l}{dU^*} = 10.612(\frac{U}{U^*})^{2}$ \\ $\frac{w}{d}\leq 50,$ $\frac{D_l}{dU^*} = (7.428+1.775(\frac{w}{d})^{0.62}(\frac{U}{U^*})^{0.572})(\frac{U}{U^*})^{2}$\end{tabular}}}        & \multicolumn{1}{c|}{\multirow{2}{*}{Propose a segmented symbolic equation}}       \\   
    \multicolumn{1}{|c|}{}   &\multicolumn{1}{|c|}{}                                                       & \multicolumn{1}{c|}{}                                                                                                                                                                                        & \multicolumn{1}{c|}{}                                    \\ [3pt]\hline
    \multicolumn{1}{|c|}{12}   &\multicolumn{1}{|c|}{Zeng and Huai / 2014}                                   & \multicolumn{1}{c|}{$\frac{D_l}{dU^*} = 5.4(\frac{w}{d})^{0.7}(\frac{U}{U^*})^{1.13}$}                                                                                                                       & \multicolumn{1}{c|}{\begin{tabular}[c]{@{}c@{}}Provide validation on \\ trapezoidal artificial channels\end{tabular}}               \\ [3pt]\hline
    \multicolumn{1}{|c|}{13}   &\multicolumn{1}{|c|}{Disley et al. / 2014}                                   & \multicolumn{1}{c|}{$\frac{D_l}{dU^*} = 3.563(\frac{U}{(gd)^{0.5}})^{-0.4117}(\frac{w}{d})^{0.6776}(\frac{U}{U^*})^{1.0132}$}                                                                                & \multicolumn{1}{c|}{Introduce a new variable into the formula}               \\ [3pt]\hline
    \end{tabular}%

    \begin{tablenotes}
        \item[*] $D_l$ - the longitudinal dispersion coefficient; $w$ - the channel width; $d$ - the channel depth;  
                 $U$ - The flow velocity; $U^*$ - The shear velocity; $S$ - the energy slope; $\epsilon$ - equal to $0.145+(\frac{(\frac{w}{d})^{1.38}(\frac{U}{U^*})}{3520})$;
                 $g$ - the acceleration of gravity. 
                
    \end{tablenotes}
    \end{threeparttable}}
    \label{T:statisticalEq}
\end{table}  

However, inconsistency has been found between predictions of those statistical models and experimental results. The performance of those equations 
could vary widely under the same stream and geometric 
conditions\cite{noori2011a, ramezani2019numerical}. That is because those models are basically regressions of 
their own training datasets. But those methods are not capable of modeling this highly nonlinear phenomenon with only hundreds of samples. 
Besides, the lack of proper data pre-processing and data scarcity introduce uncertainty into those models. 
These defects are the main reason for this inconsistency\cite{ramezani2019numerical}. 
For a more precise prediction, a modeling method 
with better general applicability for $D_l$ and improvement on data are required. 

The rise of ML-driven methods in related research provides a possible solution. Machine learning
techniques have powerful regression ability, even on biased data. Taking neural networks(NN) for example, studies 
show that a two-layer NN can represent any underlaying distribution in the data\cite{waibel1988phoneme, myles1997estimating, toulson1992data}.
Moreover, the regression ability of those methods is not only limited to interpolation but also extrapolation.
Its application potential has been verified in many scientific and engineering 
applications\cite{aguilera2011review, granata2017machine, mudunuru2021physics}.
On the prediction of $D_l$, several ML-driven studies have revealed apparent superiority over other methods, especially in accuracy and precision.
Based on whether symbolic equations will be proposed, the used ML method can be further divided into implicit and explicit methods.

Implicit ML-driven methods refer to various kinds of machine learning techniques which cannot give clear and simple symbolic expressions of models.
Representative methods are neural network (NN) and support vector machine (SVM). Those methods are generally used as \emph{black boxes} in applications.
However, It doesn't mean that such methods cannot provide symbolic equations for their calculation process. They are used as black
boxes mainly because their symbolic equations are too complicated to represent and understand. On the contrary, the explicit ML-driven method can
give a much simpler mathematical model, which is easy to use, interpret and give physical meaning, increasing 
both our predictive power and the insight behind the complex phenomenon. 

Implicit ML-driven methods are used earlier on the prediction of $D_l$. NNs with different designs were 
used to forecast the dispersion process 
by Tayfur and Singh\cite{tayfur2005predicting}, Topark and Cigizoglu\cite{toprak2008predicting}, and Noori et al.\cite{noori2011a}; 
The performance of the adaptive neuro-fuzzy inference system(ANFIS) was examined by Noori et al.\cite{noori2009predicting} and Riahi-Madvar et al.\cite{riahi-madvar2009an}; 
Azamathulla and Wu\cite{azamathulla2011support} developed a support 
vector machine(SVM) approach for this problem; Hybrid genetic expression programming(GEP) was 
applied on this topic by Alizadeh et al. \cite{alizadeh2017predicting} 
and four different metaheuristic algorithms were evaluated; The advantages and disadvantages of NN and GEP 
were discussed by Seifi and Riahi-Madvar \cite{seifi2019improving}; Ghiasi et al.\cite{ghiasi2019granular} carried out a successful application 
of the granular computing (GRC) method as well as sensitivity analysis. Results show that the parameter, $\frac{w}{d}$ matters most 
on the output in all models. Among them, NN is the most widely used 
method\cite{tayfur2005predicting, toprak2008predicting, noori2011a, alizadeh2017predicting, seifi2019improving, ghiasi2019granular}. 
All of those machine learning techniques achieve excellent predictions on $D_l$. But their black-box nature limits further
understandings of underlying physics and bring uncertainty into field application. 

To remedy the black-box attribute, several explicit ML-driven methods dedicated to finding symbolic formulas have been used. Sahay and Dutta\cite{sahay2009prediction}
reported an application of a genetic algorithm for the regression of symbolic equations of $D_l$ formulas. A concise equation was 
regressed directly from data; 
Etemad-Shahidi and Taghipour\cite{etemad-shahidi2012predicting} combined the M5 model tree and 149 data samples to propose a novel symbolic result. 
However, the performance of this symbolic model was unsatisfying; 
By minimizing the sum-square error, a differential evolution(DE) algorithm was successfully applied to this problem by Li et al.\cite{li2013differential}. 
65 samples from 29 rivers in the USA were analyzed; Satter and Gharabaghi developed two gene expression models of 
$D_l$ and its various hydraulic variables, including the Froude number, aspect ratio, and the bed material roughness\cite{sattar2015gene}.
A parametric analysis is performed for further verification; Wang et al. proposed a more concise and accurate 
canonical equation form by distilling the physical meaning of dispersion\cite{wang2016estimating}. 
Based on the novel combination, genetic programming without pre-specified correlations combined with a theoretical method was used to fit a 
physically sound explicit predictor\cite{wang2016estimating, wang2017physically}; 
A multi-objective particle swarm (PSO) optimization technique was applied to drive a novel equation to estimate the D by Alizadeh et al.\cite{alizadeh2017predicting}. 
Extensive field data, including various hydraulic and geometric parameters, were utilized; 
On the same 503-sample dataset, Riahi-Madvar et al. \cite{riahi-madvar2019pareto} and Memarzadeh, R., et al. \cite{memarzadeh2020a} 
combined optimization algorithms with the Subset Selection of Maximum Dissimilarity(SSMD), a division method for the 
training set and testing set. Results show the model of Memarzadeh, R., et al. \cite{memarzadeh2020a} has better accuracy; 
Riahi-Madvar et al.\cite{madvar2020derivation} implemented the prediction equation of $D_l$ into the NN and utilized the powerful regression capabilities of NN to fit parameters. 
However, the equation obtained was overly complicated. Explicit formulas from the above studies are shown in Table \ref{T:explicitEq}.
As can be seen, the formulas are complicated, with little physical meaning. It is reasonable to suspect that they are subject of 
overfitting.

\begin{table}[]
    \caption{Summary of explicit ML-driven formulas on $D_l$}
    \resizebox{\textwidth}{!}{
    \begin{threeparttable}
    \begin{tabular}{cccc}
    \hline
    \multicolumn{1}{|c|}{seq}   &\multicolumn{1}{|c|}{Author / Year}                                          & \multicolumn{1}{c|}{The formula}                                                                                                                                                                             & \multicolumn{1}{c|}{Application}                         \\ [3pt]\hline 
    \multicolumn{1}{|c|}{14}    &\multicolumn{1}{|c|}{Sahay and Dutta / 2009}                                 & \multicolumn{1}{c|}{$\frac{D_l}{dU^*} = 2(\frac{w}{d})^{0.96}(\frac{U}{U^*})^{1.25}$}                                                                                                                        & \multicolumn{1}{c|}{Genetic algorithm}                   \\ [3pt]\hline
    \multicolumn{1}{|c|}{15}    &\multicolumn{1}{|c|}{Li et al. / 2013}                                       & \multicolumn{1}{c|}{$\frac{D_l}{dU^*} = 2.2820(\frac{w}{d})^{0.7613}(\frac{U}{U^*})^{1.4713}$}                                                                                                               & \multicolumn{1}{c|}{Differential evolution algorithm}                 \\ [3pt]\hline
    \multicolumn{1}{|c|}{\multirow{5}{*}{16, 17}}    &\multicolumn{1}{|c|}{\multirow{5}{*}{Scatter and Gharabaghi  / 2015}}        & \multicolumn{1}{c|}{\multirow{5}{*}{\begin{tabular}[c]{@{}c@{}}$\frac{D_l}{dU^*} = a(\frac{w}{d})^{b}(\frac{U}{U^*})^{c}$ \\ $Model_{17}$: $a=2.9 \times 4.6^{\sqrt{F_r}}, b=0.5-F_r, c=1+\sqrt{F_r}, d=-0.5$ \\ $Model_{18}$: $a=8.45, b=0.5-0.514F_r^{0.516}+\frac{U}{U^*}0.42^{\frac{U}{U^*}}, c=1.65, d=0$\\ $F_r=\frac{U}{\sqrt{gd}}$\end{tabular}}}                  & \multicolumn{1}{c|}{\multirow{5}{*}{Genetic algorithm}}  \\
    \multicolumn{1}{|c|}{}      &\multicolumn{1}{|c|}{}                                                       & \multicolumn{1}{c|}{}                                                                                                                                                                                        & \multicolumn{1}{c|}{}                                    \\ 
    \multicolumn{1}{|c|}{}      &\multicolumn{1}{|c|}{}                                                       & \multicolumn{1}{c|}{}                                                                                                                                                                                        & \multicolumn{1}{c|}{}                                    \\ 
    \multicolumn{1}{|c|}{}      &\multicolumn{1}{|c|}{}                                                       & \multicolumn{1}{c|}{}                                                                                                                                                                                        & \multicolumn{1}{c|}{}                                    \\ 
    \multicolumn{1}{|c|}{}      &\multicolumn{1}{|c|}{}                                                       & \multicolumn{1}{c|}{}                                                                                                                                                                                        & \multicolumn{1}{c|}{}                                    \\ [3pt]\hline
   
    \multicolumn{1}{|c|}{18}    &\multicolumn{1}{|c|}{Wang and Huai / 2016}                                   & \multicolumn{1}{c|}{$\frac{D_l}{dU^*} = 17.648(\frac{w}{d})^{0.3619}(\frac{U}{U^*})^{1.16}$}                                                                                                                & \multicolumn{1}{c|}{Genetic algorithm}                    \\ [3pt]\hline
    \multicolumn{1}{|c|}{19}    &\multicolumn{1}{|c|}{Wang and Huai / 2017}                                   & \multicolumn{1}{c|}{$\frac{D_l}{dU^*} = (0.718+47.9\frac{d}{w})\frac{U}{w}$}                                                                                                                                & \multicolumn{1}{c|}{Genetic algorithm}                    \\ [3pt]\hline
    \multicolumn{1}{|c|}{\multirow{2}{*}{20}}    &\multicolumn{1}{|c|}{\multirow{2}{*}{Alizadeh et al. / 2017}}                & \multicolumn{1}{c|}{\multirow{2}{*}{\begin{tabular}[c]{@{}c@{}}$\frac{w}{d}>28,$ $\frac{D_l}{dU^*} = 9.931(\frac{w}{d})^{0.187}(\frac{U}{U^*})^{1.802}$ \\ $\frac{w}{d}\leq 28,$ $\frac{D_l}{dU^*} = 5.319(\frac{w}{d})^{1.206}(\frac{U}{U^*})^{0.075}$\end{tabular}}}        & \multicolumn{1}{c|}{\multirow{2}{*}{Optimization algorithm}}   \\   
    \multicolumn{1}{|c|}{}      &\multicolumn{1}{|c|}{}                                                       & \multicolumn{1}{c|}{}                                                                                                                                                                                        & \multicolumn{1}{c|}{}                                    \\ [3pt]\hline
       
    \multicolumn{1}{|c|}{\multirow{3}{*}{21}}    &\multicolumn{1}{|c|}{\multirow{3}{*}{Riahi-Madvar et al. / 2019}}            & \multicolumn{1}{c|}{\multirow{3}{*}{\begin{tabular}[c]{@{}c@{}} $\frac{D_l}{dU^*}=33.99(\frac{w}{d})^{0.5}+8.497\frac{w}{d}(\frac{U^*}{U})^2+\frac{8.497wU^*}{dU}$ \\ $16.99\frac{wU^*}{dU}+\frac{0.0000486(\frac{w}{d})^{0.5}-0.00021}{d^{1.5}(U^*)^{4}}w^{1.6}U^4+0.01478$ \end{tabular}}}       & \multicolumn{1}{c|}{\multirow{3}{*}{Optimization algorithm}}  \\   
    \multicolumn{1}{|c|}{}      &\multicolumn{1}{|c|}{}                                                       & \multicolumn{1}{c|}{}                                                                                                                                                                                        & \multicolumn{1}{c|}{}                                     \\ 
    \multicolumn{1}{|c|}{}      &\multicolumn{1}{|c|}{}                                                       & \multicolumn{1}{c|}{}                                                                                                                                                                                        & \multicolumn{1}{c|}{}                                     \\ [3pt]\hline

    \multicolumn{1}{|c|}{\multirow{4}{*}{22,23}}    &\multicolumn{1}{|c|}{\multirow{4}{*}{Memarzadeh, R., et al. / 2020}}            & \multicolumn{1}{c|}{\multirow{4}{*}{\begin{tabular}[c]{@{}c@{}} $Model_{22}$: $\frac{w}{d}>27, \frac{D_l}{dU^*}=(0.35+8.7(\frac{d}{w}))(6.4+8(\frac{w}{d}))(\frac{U}{U^*})^{0.5}$\\ $\frac{w}{d}\leq 27, \frac{D_l}{dU^*}=0.2694(\frac{w}{d})^{2.2456}$\\ $Model_{23}$: For all data, $\frac{D_l}{dU^*} = 4.5(\frac{w}{d})(\frac{U}{U^*})^{0.5}$ \end{tabular}}}       & \multicolumn{1}{c|}{\multirow{4}{*}{Optimization algorithm}}  \\   
    \multicolumn{1}{|c|}{}      &\multicolumn{1}{|c|}{}                                                       & \multicolumn{1}{c|}{}                                                                                                                                                                                        & \multicolumn{1}{c|}{}                                     \\ 
    \multicolumn{1}{|c|}{}      &\multicolumn{1}{|c|}{}                                                       & \multicolumn{1}{c|}{}                                                                                                                                                                                        & \multicolumn{1}{c|}{}                                     \\ 
    \multicolumn{1}{|c|}{}      &\multicolumn{1}{|c|}{}                                                       & \multicolumn{1}{c|}{}                                                                                                                                                                                        & \multicolumn{1}{c|}{}                                     \\ [3pt]\hline

    \multicolumn{1}{|c|}{\multirow{15}{*}{24,25}}    &\multicolumn{1}{|c|}{\multirow{15}{*}{Riahi-Madvar et al. / 2020}}           & \multicolumn{1}{c|}{\multirow{15}{*}{\begin{tabular}[c]{@{}c@{}} $Model_{24}$: \\ $a = 1 + e^{-0.02w+0.39d+3.52U+11.37U^*-3.72}$\\ $b = 1 + e^{0.02w-0.48d+0.69U+11.37U^*+2.37}$ \\ $c = 1 + e^{0.02w+0.87d-3.52U-2.04U^*-4.48}$ \\ $d = 1 + e^{0.03w+1.6d+3.52U-4.49U^*-11.6}$\\ $D_l=\frac{-124.74}{a}+\frac{374.99}{b}-\frac{517.15}{c}-\frac{636.76}{d}+227.59$ \\ \\ $Model_{25}$: \\$a=1+e^{0.04w-0.62d-2.71U+23.26U^*-9.21}$\\ $b=1+e^{-0.023w+1.31d+0.54U+10.18U^*+1.91}$ \\$c=1+e^{0.021w+0.11d+2.04U-3.60U^*-7.25}$ \\$d=1+e^{0.01w+1.07d+2.14U+0.335U^\ast-7.20}$ \\$e=1+e^{-0.01w-0.24d+7.94U+1.49U^\ast+2.33}$ \\ $D_l=\frac{471.22}{a}+\frac{315.96}{b}-\frac{306.77}{c}-\frac{818.23}{d}-\frac{583.71}{e}+227.59$ \end{tabular}}}       & \multicolumn{1}{c|}{\multirow{15}{*}{Neural network}}  \\   
    \multicolumn{1}{|c|}{}      &\multicolumn{1}{|c|}{}                                                       & \multicolumn{1}{c|}{}                                                                                                                                                                                        & \multicolumn{1}{c|}{}                                     \\ 
    \multicolumn{1}{|c|}{}      &\multicolumn{1}{|c|}{}                                                       & \multicolumn{1}{c|}{}                                                                                                                                                                                        & \multicolumn{1}{c|}{}                                     \\ 
    \multicolumn{1}{|c|}{}      &\multicolumn{1}{|c|}{}                                                       & \multicolumn{1}{c|}{}                                                                                                                                                                                        & \multicolumn{1}{c|}{}                                     \\ 
    \multicolumn{1}{|c|}{}      &\multicolumn{1}{|c|}{}                                                       & \multicolumn{1}{c|}{}                                                                                                                                                                                        & \multicolumn{1}{c|}{}                                     \\ 
    \multicolumn{1}{|c|}{}      &\multicolumn{1}{|c|}{}                                                       & \multicolumn{1}{c|}{}                                                                                                                                                                                        & \multicolumn{1}{c|}{}                                     \\ 
    \multicolumn{1}{|c|}{}      &\multicolumn{1}{|c|}{}                                                       & \multicolumn{1}{c|}{}                                                                                                                                                                                        & \multicolumn{1}{c|}{}                                     \\ 
    \multicolumn{1}{|c|}{}      &\multicolumn{1}{|c|}{}                                                       & \multicolumn{1}{c|}{}                                                                                                                                                                                        & \multicolumn{1}{c|}{}                                     \\ 
    \multicolumn{1}{|c|}{}      &\multicolumn{1}{|c|}{}                                                       & \multicolumn{1}{c|}{}                                                                                                                                                                                        & \multicolumn{1}{c|}{}                                     \\ 
    \multicolumn{1}{|c|}{}      &\multicolumn{1}{|c|}{}                                                       & \multicolumn{1}{c|}{}                                                                                                                                                                                        & \multicolumn{1}{c|}{}                                     \\ 
    \multicolumn{1}{|c|}{}      &\multicolumn{1}{|c|}{}                                                       & \multicolumn{1}{c|}{}                                                                                                                                                                                        & \multicolumn{1}{c|}{}                                     \\ 
    \multicolumn{1}{|c|}{}      &\multicolumn{1}{|c|}{}                                                       & \multicolumn{1}{c|}{}                                                                                                                                                                                        & \multicolumn{1}{c|}{}                                     \\ 
    \multicolumn{1}{|c|}{}      &\multicolumn{1}{|c|}{}                                                       & \multicolumn{1}{c|}{}                                                                                                                                                                                        & \multicolumn{1}{c|}{}                                     \\ 
    \multicolumn{1}{|c|}{}      &\multicolumn{1}{|c|}{}                                                       & \multicolumn{1}{c|}{}                                                                                                                                                                                        & \multicolumn{1}{c|}{}                                     \\ 
    \multicolumn{1}{|c|}{}      &\multicolumn{1}{|c|}{}                                                       & \multicolumn{1}{c|}{}                                                                                                                                                                                        & \multicolumn{1}{c|}{}                                     \\ [3pt]\hline
    \end{tabular}%
    \begin{tablenotes}
        \item[*] $D_l$ - the longitudinal dispersion coefficient; $w$ - the channel width; $d$ - the channel depth;  
                 $U$ - The flow velocity; $U^*$ - The shear velocity; $F_r$ - Froude number.              
    \end{tablenotes}
    \end{threeparttable}}
    \label{T:explicitEq}
\end{table}

Nevertheless, those explicit ML-driven methods are weaker than implicit methods in 
performance\cite{riahi-madvar2009an, sahay2009prediction, sattar2015gene,memarzadeh2020a}
Detailed Comparisons can be found in Section \ref{sec:evaluation}. 
This is generally caused by two factors: data scarcity and algorithm defects. Most of the studies 
use only dozens of data, which is far from enough to predict this highly nonlinear process. 
More data samples are needed. 

Apart from the data problem, those algorithms suffer from several defects. Most of the algorithms 
are doing fitting for parameters of a 
fixed formula form(Eq. \ref{eq:generalForm}). 

\begin{equation}
    D_l = p_1 \left(\frac{w}{d} \right)^{p_2} \left(\frac{U}{U^*} \right)^{p_3}
    \label{eq:generalForm} 
\end{equation}
where $p_1$, $p_2$ and $p_3$ are different parameters.

Few studies have explored other possible combinations of functional forms and parameters. Some of them do 
try different symbolic equation forms, but mainly by test and 
trail\cite{sattar2015gene, madvar2020derivation} which is inefficient and incomplete. Besides, the 
main drawback of those algorithms is extrapolation ability. 
These models perform poorly under new data samples (See details in Section \ref{sec:evaluation}). Those factors 
make the present explicit formulae unsuitable
to satisfy the application demand on the prediction of $D_l$. Due to these reasons, there is a pressing need 
for an improved symbolic regression framework. 
    
To propose a high-performance, interpretable $D_l$ prediction equation, the above two obstacles need 
to be tackled correspondingly. For data scarcity, 
a large and convincing dataset is needed. For algorithm defects, a novel scheme is required. On one hand, 
It should be able to explore more 
possible formula forms exhaustively and find the more accurate combination. On the other hand, it needs to 
have strong regression ability, which can achieve 
better results in interpolation and extrapolation. Details on solutions for these two obstacles will be discussed in
Section \ref{sec:data} and Section \ref{sec:method} separately.

\section{Data}\label{sec:data}
\subsection{Data exploratory and pre-process}\label{sec:exploratoryAna}
Data is pivotal to the development of governing equations. Generally, samples and features of the used dataset 
will determine the upper limit of the predictive model. 
Different regression methods are just struggling to reach this limit\cite{rahm2000data}. Therefore, it is 
necessary to do some analysis and processing 
on the data foundation.

In this study, a dataset of 1094 samples, which contain measured $D_l$ and its influence variables ($w, d, U$, and $U^*$)
in various streams all over the world, is collected from 
published works\cite{seo1998predicting, ZhiQiangDeng.2001, S.M.Kashefipour.2002, 
GokmenTayfur.2005, MeredithL.Carr.2007, 
Z.FuatToprak.2008,RoohollahNoori.2009, AmirEtemadShahidi.2012, 
RasoulMemarzadeh.2020}.

Since these samples are collected from different streams globally, it is helpful to have preliminary knowledge 
of the patterns behind them. According to the adversarial validation test(Presented in supplementary material, Section S1),
there are three groups of samples, which demonstrates the data diversity.



\begin{figure}[]
    \begin{centering}
    \includegraphics[width=0.9\linewidth]{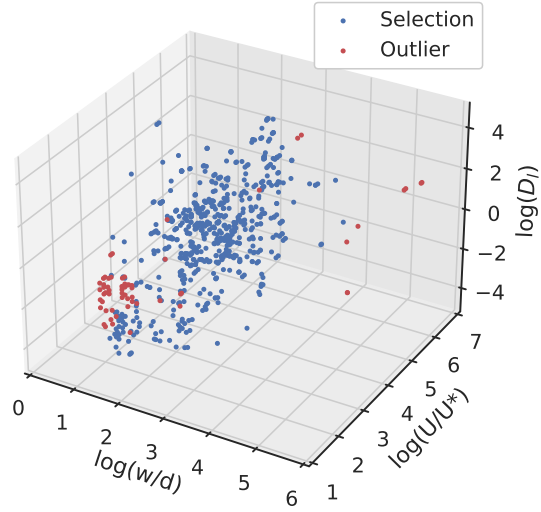}
    \caption{The visulization of selections and outliers}
    \label{fig:outlier_cleaning}
    \end{centering}
\end{figure}

After deleting null values and duplication, a dataset of 721 samples is obtained. But there are some outliers in the data(Fig. \ref{fig:outlier_cleaning}). 
Many previous studies have mentioned the influence of extreme values\cite{ alizadeh2017improvement, sahay2011prediction, S.M.Kashefipour.2002}. 
This effect is usually solved through removal by tests and trails, 
which has low efficiency and lack of theoretical support. 
Such extreme values are highly possible to be outliers caused by variability in measurement or experimental errors\cite{hand2014data}. 
The existing of them can bring uncertainty into the result.
But few previous studies conducted reasonable outlier removal. 
To remove those errors, the Inter Quartile Range(IQR) is introduced. IQR is a measure of statistical 
discrete degree (Eq. \ref{eq:iqr}).
It can reveal the dispersion of variables in the dataset, which is often used in removal of the 
outlier. 
\begin{equation}
    IQR = Q3 - Q1
    \label{eq:iqr}
\end{equation}
where for a $2n$ or $2n+1$ set of samples, $Q3$ = the median of the n largest samples; $Q1$ = the median of the n smallest samples. 

Samples out of [$Q1 - 1.5*IQR$, $Q3 + 1.5*IQR$] are usually considered as 
outliers. After selection by IQR, a dataset of 660 samples is obtained 
(Fig. \ref{fig:outlier_cleaning}).  

\begin{figure}[]
    \centering
    \subfigure[the channel width]{
    \begin{minipage}[t]{0.15\linewidth}
    \centering
    \includegraphics[width=1in]{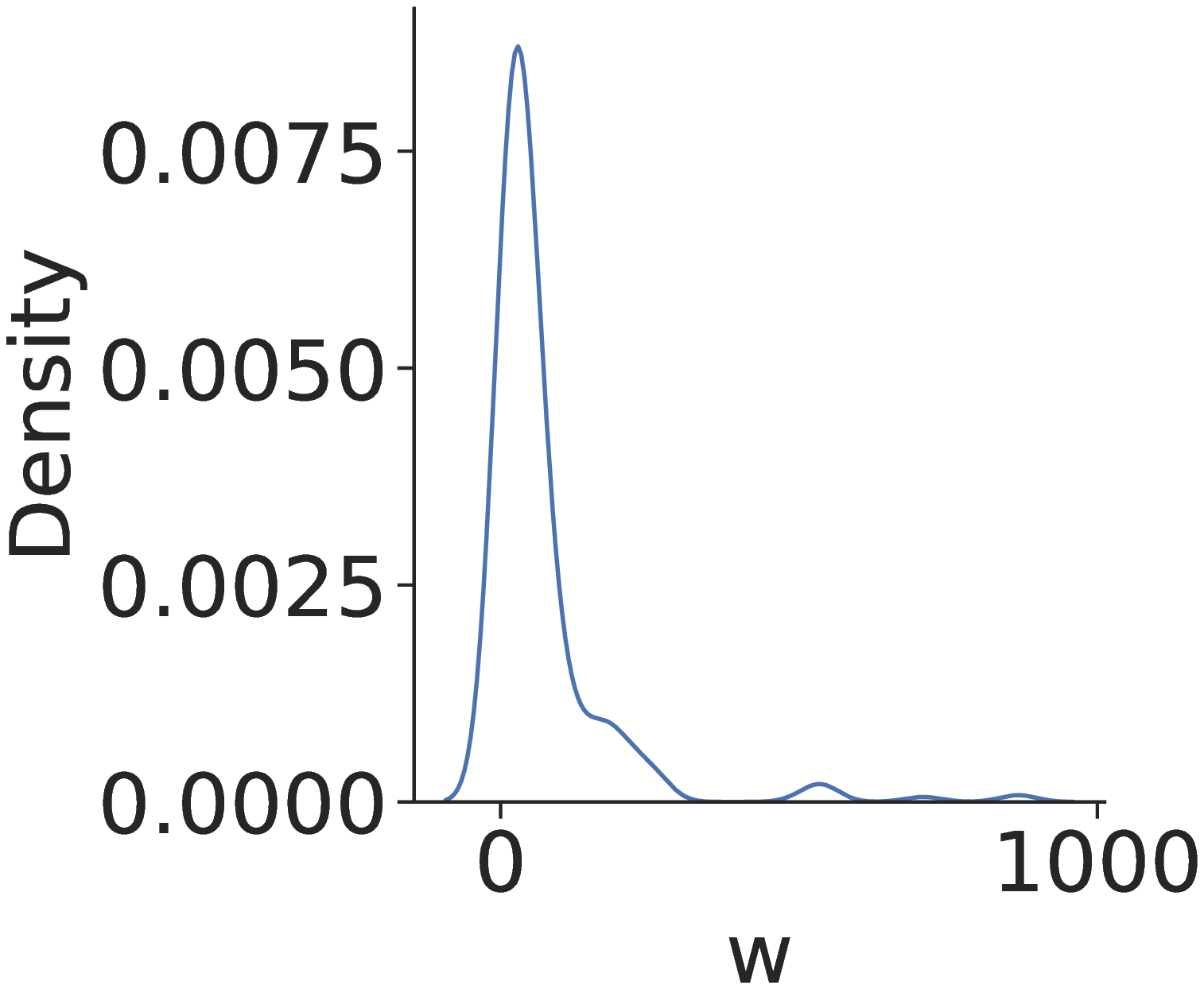}
    \end{minipage}%
    }%
    \subfigure[the channel depth]{
    \begin{minipage}[t]{0.15\linewidth}
    \centering
    \includegraphics[width=1in]{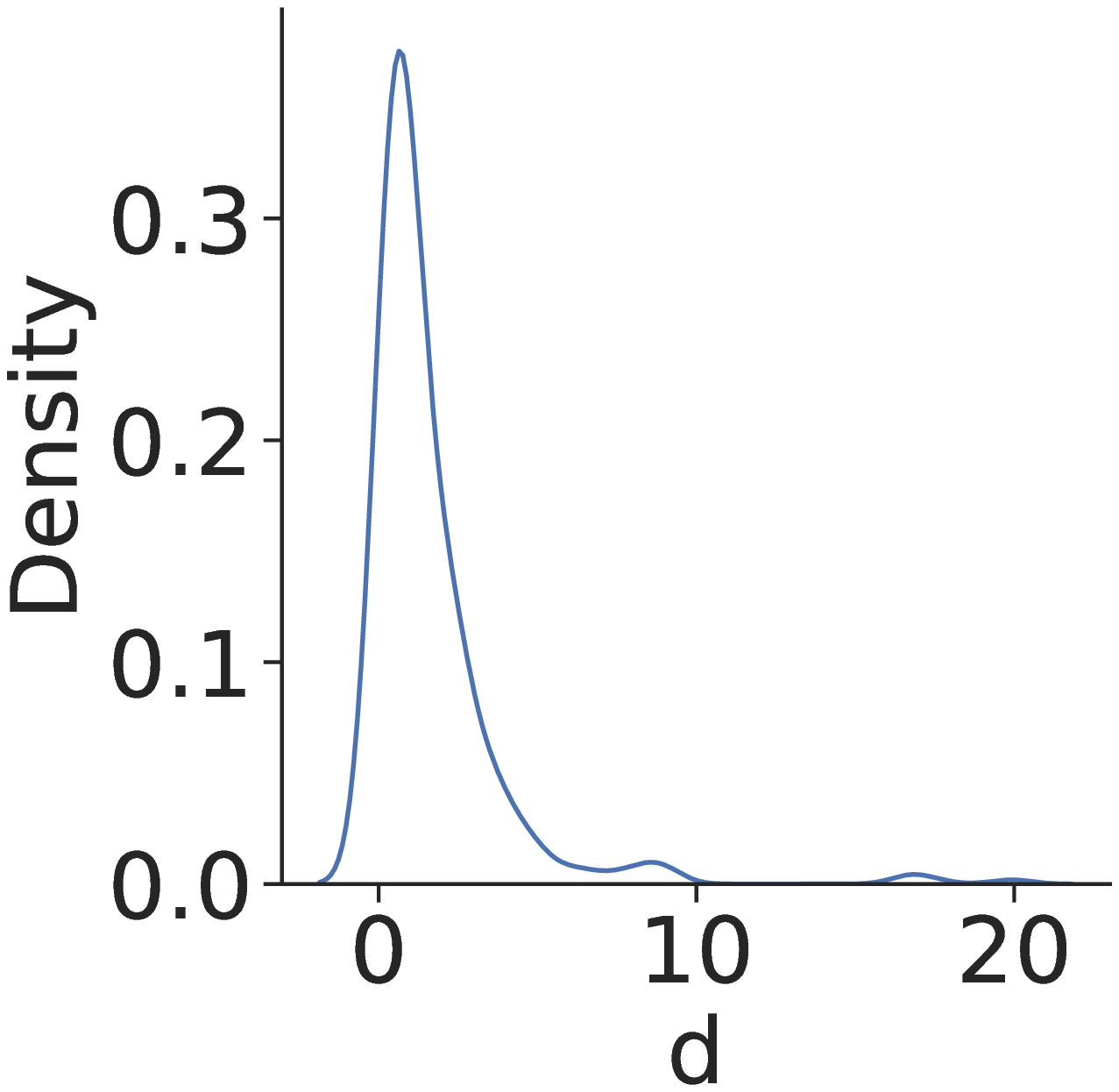}
    \end{minipage}%
    }%
    \subfigure[the flow velocity]{
    \begin{minipage}[t]{0.15\linewidth}
    \centering
    \includegraphics[width=1in]{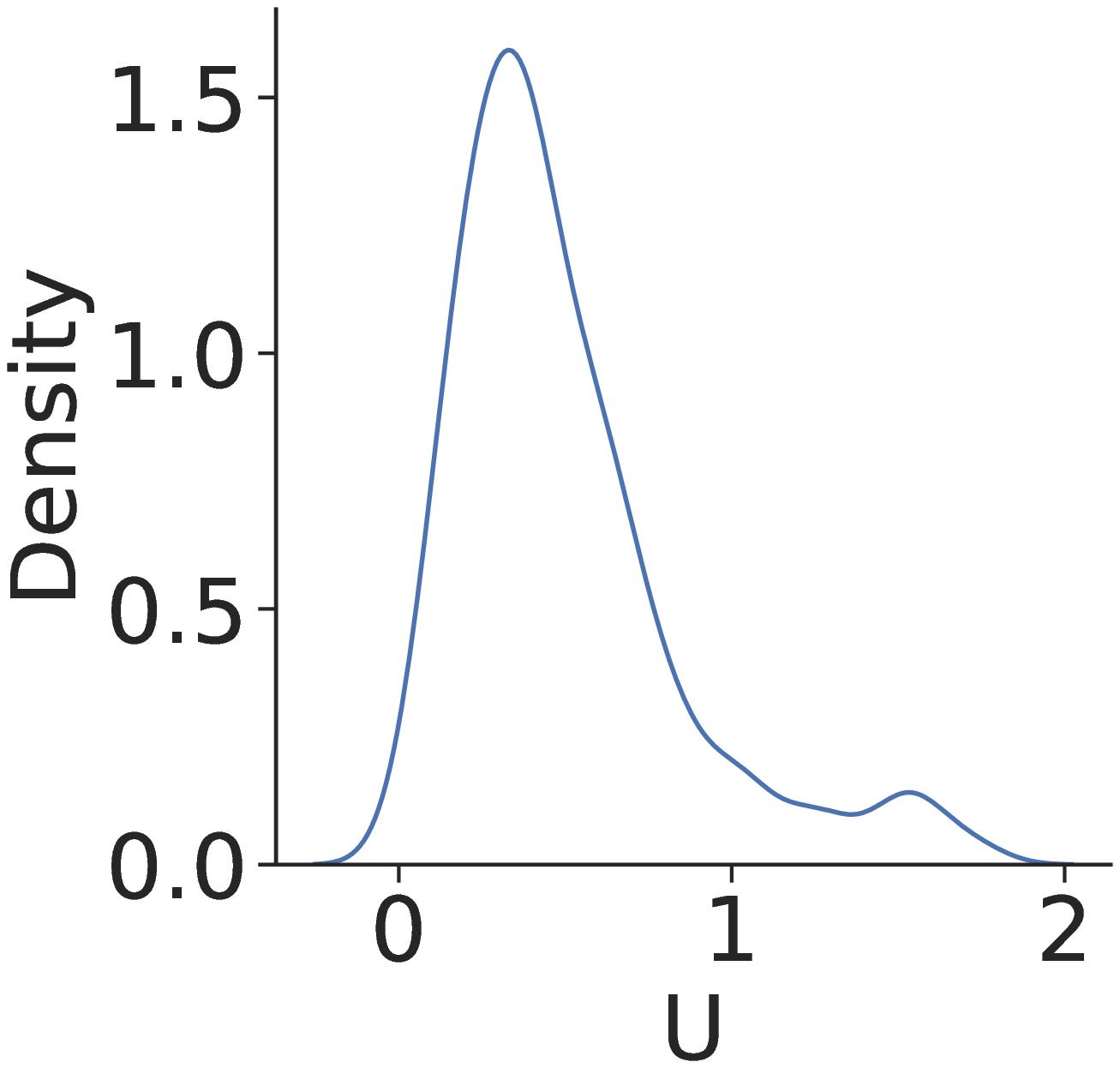}
    \end{minipage}
    }%
    \subfigure[the shear velocity]{
    \begin{minipage}[t]{0.15\linewidth}
    \centering
    \includegraphics[width=1in]{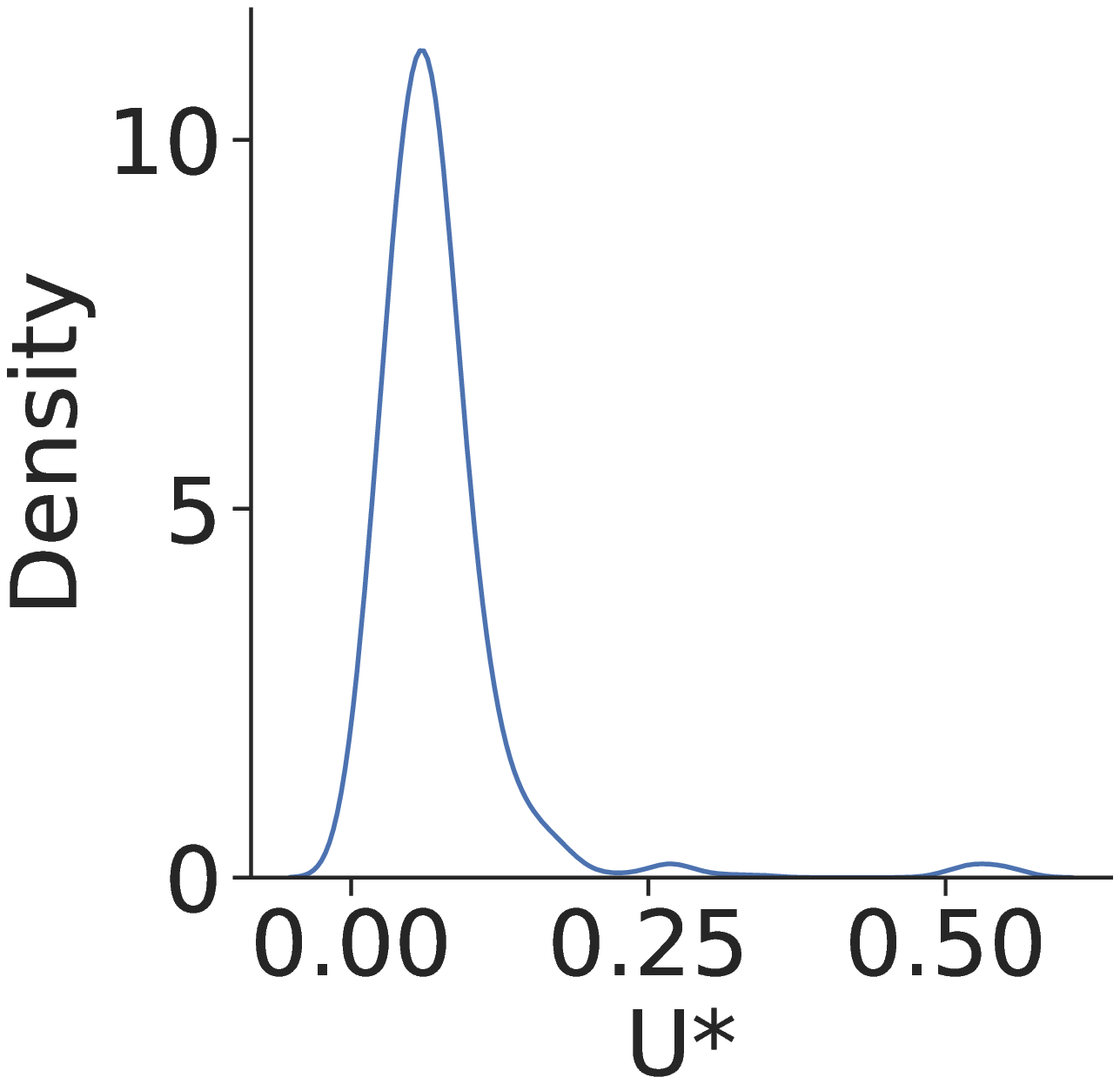}
    \end{minipage}}
    \subfigure[the $D_l$]{
        \begin{minipage}[t]{0.15\linewidth}
        \centering
        \includegraphics[width=1in]{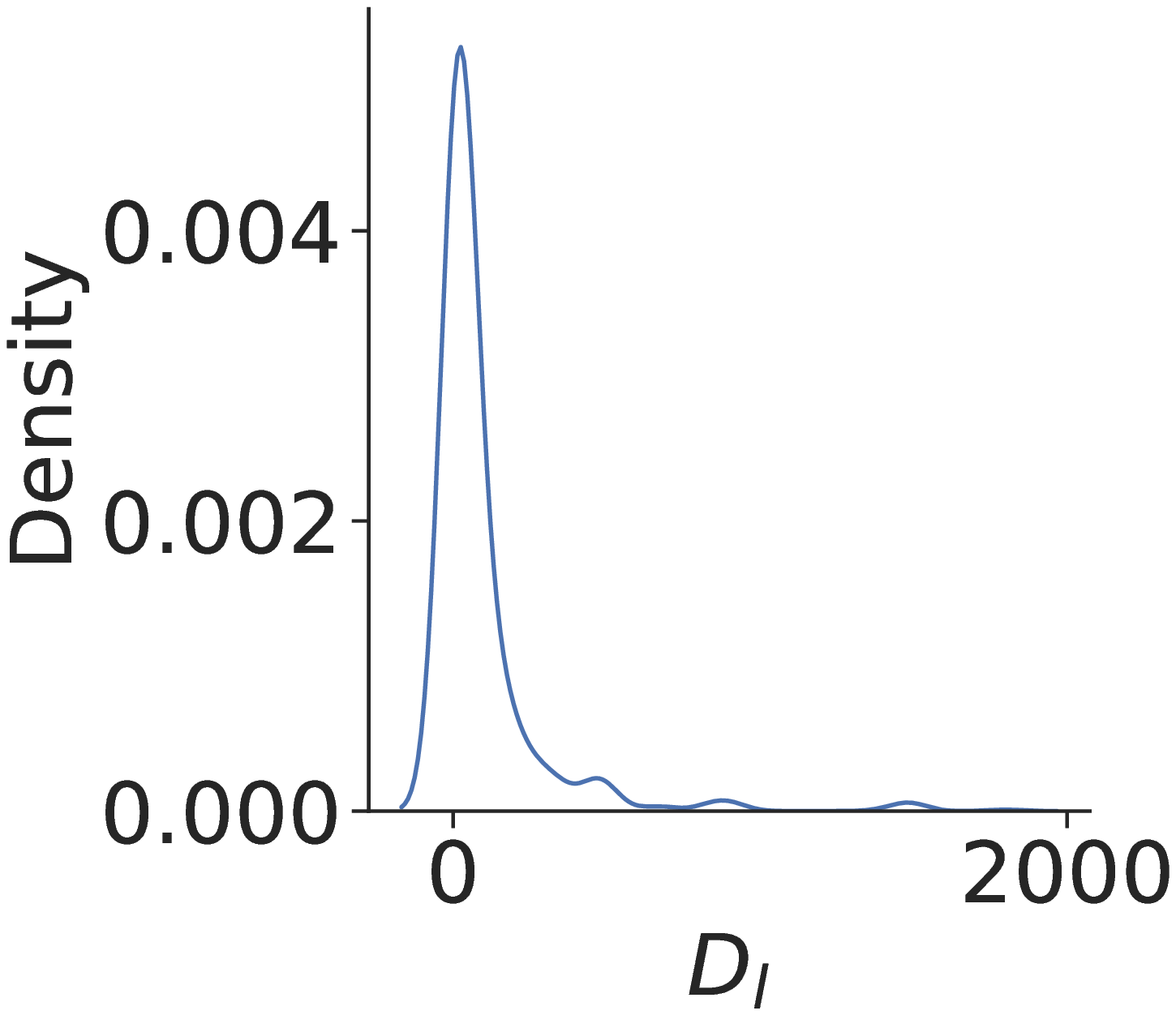}
    \end{minipage}}    
    \centering
    \caption{The kenerl density plot of four input parameters}
    \label{fig:displot}
\end{figure}

\begin{figure}[]
    \begin{centering}
    \includegraphics[width=0.9\linewidth]{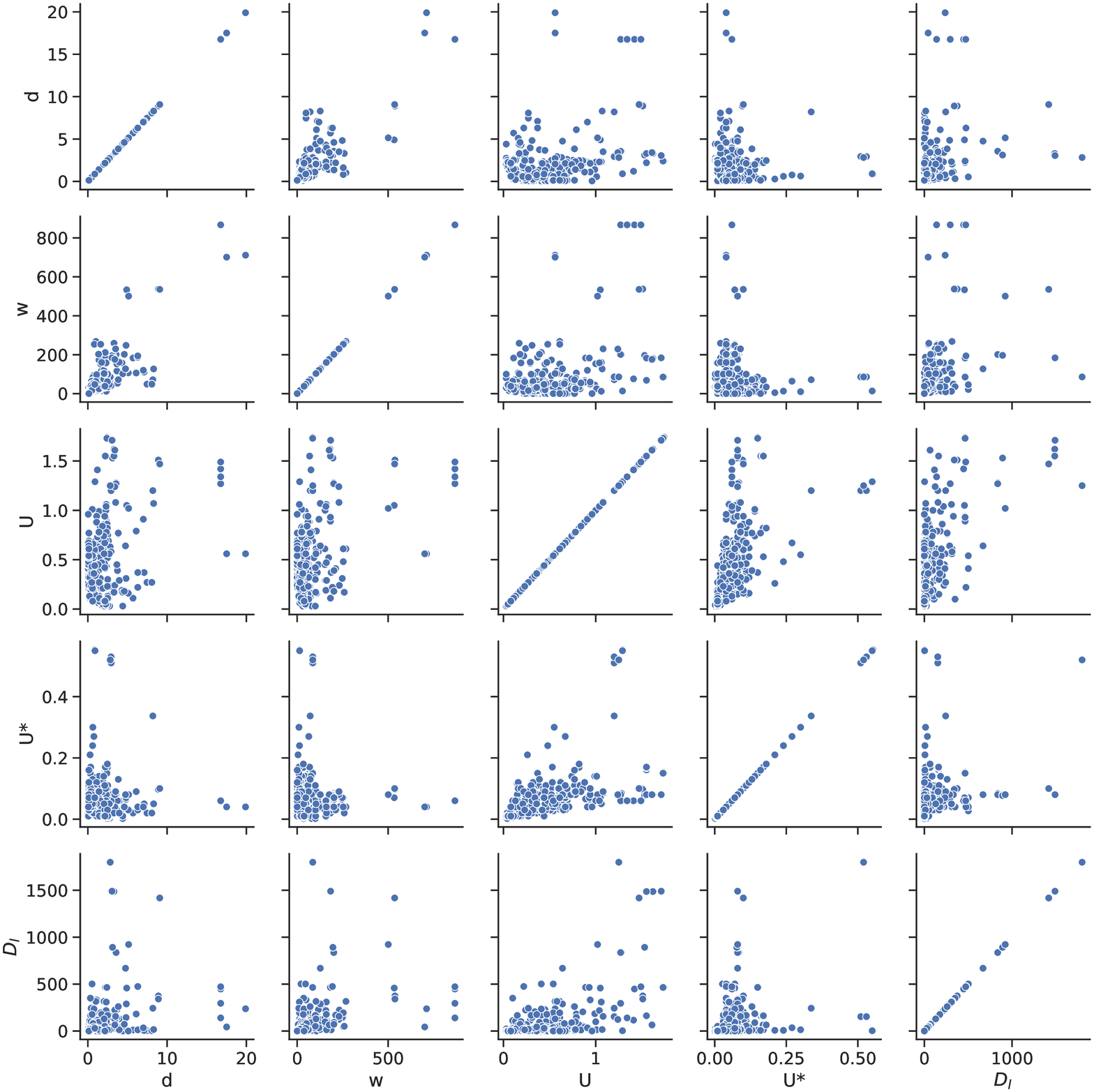}
    \caption{The pairplot of the data after cleaning}
    \label{fig:Ppairplot_no}
    \end{centering}
\end{figure}

To have a basic understanding of the data distribution in the processed dataset, the kenerl density plot (KDP, Fig. \ref{fig:displot}) 
and the pair plot (Fig. \ref{fig:Ppairplot_no}) are carried out.
KDE aims to reveal the underlying distribution density of samples with the Gaussian 
kernel\cite{chung2020gaussian}. It shows that all parameters in this dataset have a normal distribution, which 
satisfies the Central Limit Theorem. This will ensure the extraction formula to be universal and robust\cite{hand2014data}.
The pairplot can reveal the pairwise relationships between every two involved parameters in a group. 
As shown in Fig. \ref{fig:Ppairplot_no}, the relationship between the input parameters ($w$, $d$, $U$ and $U^*$) and the output 
parameter ($D_l$) is not linear and hard to get useful information.

\begin{figure}[]
    \begin{centering}
    \includegraphics[width=0.4\linewidth]{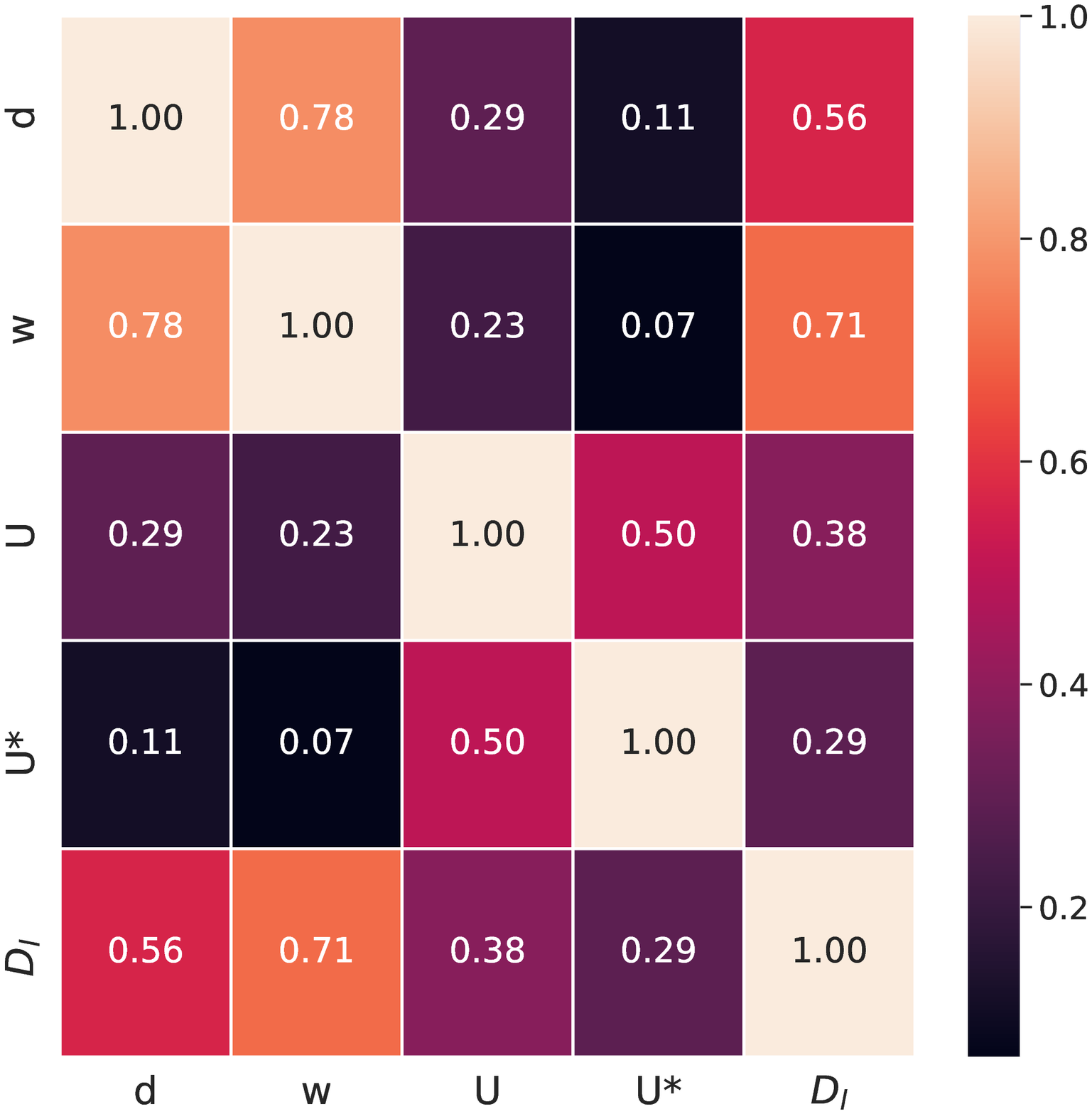}
    \caption{The SCC plot of $w$, $d$, $U$ and $U^*$}
    \label{fig:SCC}
    \end{centering}
\end{figure}

Therefore, the Spearman Correlation Coefficient (SCC) is calculated. SCC is a statistical 
measure of the monotonic correlation between variables\cite{asuero2006correlation}. 
It is less affected by outliers than Pearson Correlation Coefficient, which is used in 
many studies\cite{seo1998predicting,alizadeh2017improvement,IlWonSeo.1998}
to reveal parameter relationships. Fig. \ref{fig:SCC} is the SCC plot of input and output parameters, in which the number 
stands for the SCC value between the two involved parameters.

The Figure shows that channel geometries have stronger relationships with $D_l$ than stream 
properties ($D_l-d$=0.56, $D_l-w=0.71 > D_l-U=0.38$, $D_l-U_s$=0.29). In 
channel geometries, the channel width, $w$ has the strongest influence, which means $w$ is 
the most important variable to $D_l$. 
It is worth noting that stream properties have a weak relationship with channel 
geometrics, especially $U_s$ ($U_s-w$=0.07, $U_s-d$=0.09). 
Besides, 
channel geometrics and streams properties both have a strong internal 
correlation ($w-d$=0.78, $U-U^*$=0.50), verifying the rationality 
of data.

The statistical information on this dataset is shown in Table \ref{T:statOfData}. Selected 
metrics include: number (Num), minimum (Min), median (Med), maximum (Max), interquartile 
range (IQR), standard deviation (std), variance (Var), kurtosis (Kurt), median absolute 
deviation (MSD), and skewness(Skew).

\begin{table}[]
    \caption{The statistical analysis of the data after cleaning}
    \resizebox{\textwidth}{!}{
    \begin{tabular}{|c|c|c|c|c|c|c|c|c|c|c|}
    \cline{1-11}
    Parameter & Num             & Min               & Med                & Max             & IQR                 & std              & Var             & Kurt          & MAD          & Skew       \\ \cline{1-11}
    d         & 660             & 0.03              & 0.88               & 19.94           & 1.61                & 2.31             & 5.33            & 25.72         & 0.53         & 4.39       \\ \cline{1-11}
    w         & 660             & 0.20              & 34.95              & 867             & 47.75               & 113.06           & 12783           & 21.95         & 19.95        & 4.26       \\ \cline{1-11}
    U         & 660             & 0.03              & 0.41               & 1.74            & 0.36                & 0.35             & 0.12            & 2.26          & 0.18         & 1.52       \\ \cline{1-11}
    U*        & 660             & 0.002             & 0.06               & 0.553           & 0.040               & 0.065            & 0.004           & 29.009        & 0.020        & 4.691      \\ \cline{1-11}
    D         & 660             & 0.01              & 25.90              & 1798.60         & 62.25               & 205.59           & 42265           & 26.31         & 20.55        & 4.63       \\ \cline{1-11}
    \end{tabular}}
    \label{T:statOfData}
\end{table}

\subsection{Training and testing sets}\label{sec:TrainAndTest}
For the regression of $D_l$ equation, 462 samples (about 70\% of the overall dataset) will be used for development 
and the remaining 198 samples (30\% of the overall dataset) will be used for testing. 

Testing set selection is essential to model development. It can both direct the adjustment of the model's super-parameters 
and assess the performance of the model under new data\cite{miller2002subset, barz2020do}. The dataset 
used in this paper is much larger (at least 20\%) than other related research. 
And due to the diversity of samples and the satisfaction of Central Limit Theorem, the dataset also has advantages in comprehensiveness. 
Newly measured data is highly 
probable to locate within the distribution and range of our dataset. However, a randomly divided test set 
is highly possible to have distribution different from the original dataset, which will make the developed 
model biased and the testing result doubtful. 
To generate reliable subsets, the Subset Selection of Maximum Dissimilarity (SSMD) algorithm is implemented.

SSMD is a selection strategy, which dedicates to filter out samples with the greatest similarity to the original dataset\cite{kennard1969computer}. 
The details of this algorithm are presented in supplementary material, Section S2.

Fig. \ref{fig:SSMD} is the visualization of training and testing sets separated by SSMD.
 The statistics of parameters in the training and testing 
dataset are given in Table \ref{T:SSMD} . Results show that all parameters have almost the 
same distribution and characteristics in both sets. 
It is worth mentioning that the ranges of parameters are not precisely the same in training and testing sets. 
Actually, the training-set range 
contains the testing-set range and those testing samples are evenly scattered in the region of the training set. 
This can avoid the 
risk of overfitting and guarantee the robustness of the developed 
model\cite{roberts2017cross}.  

\begin{figure}[]
    \begin{centering}
    \includegraphics[width=0.5\linewidth]{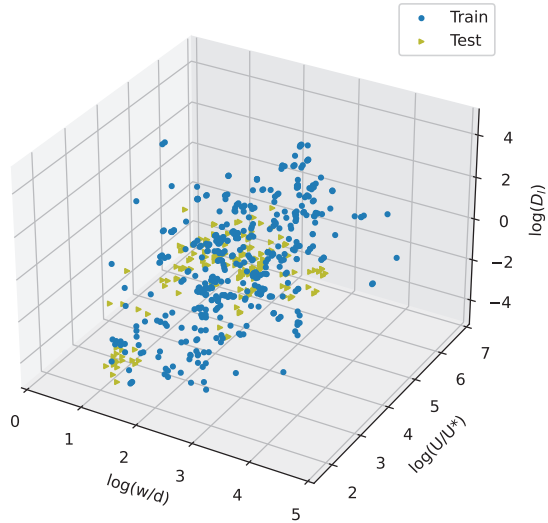}
    \caption{The visulization of training and testing sets separated by SSMD}
    \label{fig:SSMD}
    \end{centering}
\end{figure}

\begin{table}[]
    \caption{The statistical properties of training set and testing set}\centering
    \resizebox{\textwidth}{!}{
    \begin{tabular}{ccccccccccccc}
    \cline{1-11}
    \multicolumn{1}{|c|}{Subset}                        & \multicolumn{1}{c|}{Parameter} & \multicolumn{1}{c|}{Num.} & \multicolumn{1}{c|}{Min}   & \multicolumn{1}{c|}{Max}     & \multicolumn{1}{c|}{IQR}    & \multicolumn{1}{c|}{std}    & \multicolumn{1}{c|}{Var}      & \multicolumn{1}{c|}{Kurt}   & \multicolumn{1}{c|}{MAD}   & \multicolumn{1}{c|}{Skew}  &  &  \\ \cline{1-11}
    \multicolumn{1}{|c|}{\multirow{5}{*}{Training set}} & \multicolumn{1}{c|}{d}         & \multicolumn{1}{c|}{462}  & \multicolumn{1}{c|}{0.03}  & \multicolumn{1}{c|}{19.94}   & \multicolumn{1}{c|}{1.96}   & \multicolumn{1}{c|}{2.65}   & \multicolumn{1}{c|}{7.04}     & \multicolumn{1}{c|}{18.34}  & \multicolumn{1}{c|}{0.91}  & \multicolumn{1}{c|}{3.73}  &  &  \\ \cline{2-11}
    \multicolumn{1}{|c|}{}                              & \multicolumn{1}{c|}{w}         & \multicolumn{1}{c|}{462}  & \multicolumn{1}{c|}{0.20}  & \multicolumn{1}{c|}{867}     & \multicolumn{1}{c|}{73.10}  & \multicolumn{1}{c|}{130.87} & \multicolumn{1}{c|}{17128.09} & \multicolumn{1}{c|}{14.98}  & \multicolumn{1}{c|}{25.16} & \multicolumn{1}{c|}{3.56}  &  &  \\ \cline{2-11}
    \multicolumn{1}{|c|}{}                              & \multicolumn{1}{c|}{U}         & \multicolumn{1}{c|}{462}  & \multicolumn{1}{c|}{0.03}  & \multicolumn{1}{c|}{1.74}    & \multicolumn{1}{c|}{0.52}   & \multicolumn{1}{c|}{0.41}   & \multicolumn{1}{c|}{0.17}     & \multicolumn{1}{c|}{0.59}   & \multicolumn{1}{c|}{0.23}  & \multicolumn{1}{c|}{1.13}  &  &  \\ \cline{2-11}
    \multicolumn{1}{|c|}{}                              & \multicolumn{1}{c|}{U*}        & \multicolumn{1}{c|}{462}  & \multicolumn{1}{c|}{0.002} & \multicolumn{1}{c|}{0.553}   & \multicolumn{1}{c|}{0.041}  & \multicolumn{1}{c|}{0.077}  & \multicolumn{1}{c|}{0.006}    & \multicolumn{1}{c|}{20.430} & \multicolumn{1}{c|}{0.023} & \multicolumn{1}{c|}{4.044} &  &  \\ \cline{2-11}
    \multicolumn{1}{|c|}{}                              & \multicolumn{1}{c|}{$D_l$}         & \multicolumn{1}{c|}{462}  & \multicolumn{1}{c|}{0.01}  & \multicolumn{1}{c|}{1798.60} & \multicolumn{1}{c|}{121.40} & \multicolumn{1}{c|}{239.74} & \multicolumn{1}{c|}{57477.19} & \multicolumn{1}{c|}{17.96}  & \multicolumn{1}{c|}{27.45} & \multicolumn{1}{c|}{3.87}  &  &  \\ \cline{1-11}
    \multicolumn{1}{|c|}{\multirow{5}{*}{Testing set}}  & \multicolumn{1}{c|}{d}         & \multicolumn{1}{c|}{198}  & \multicolumn{1}{c|}{0.08}  & \multicolumn{1}{c|}{2.49}    & \multicolumn{1}{c|}{0.51}   & \multicolumn{1}{c|}{0.51}   & \multicolumn{1}{c|}{0.26}     & \multicolumn{1}{c|}{2.84}   & \multicolumn{1}{c|}{0.27}  & \multicolumn{1}{c|}{1.43}  &  &  \\ \cline{2-11}
    \multicolumn{1}{|c|}{}                              & \multicolumn{1}{c|}{w}         & \multicolumn{1}{c|}{198}  & \multicolumn{1}{c|}{0.40}  & \multicolumn{1}{c|}{97.54}   & \multicolumn{1}{c|}{25.83}  & \multicolumn{1}{c|}{21.60}  & \multicolumn{1}{c|}{466.53}   & \multicolumn{1}{c|}{1.04}   & \multicolumn{1}{c|}{11.23} & \multicolumn{1}{c|}{0.99}  &  &  \\ \cline{2-11}
    \multicolumn{1}{|c|}{}                              & \multicolumn{1}{c|}{U}         & \multicolumn{1}{c|}{198}  & \multicolumn{1}{c|}{0.28}  & \multicolumn{1}{c|}{0.63}    & \multicolumn{1}{c|}{0.13}   & \multicolumn{1}{c|}{0.09}   & \multicolumn{1}{c|}{0.01}     & \multicolumn{1}{c|}{0.52}   & \multicolumn{1}{c|}{0.06}  & \multicolumn{1}{c|}{0.56}  &  &  \\ \cline{2-11}
    \multicolumn{1}{|c|}{}                              & \multicolumn{1}{c|}{U*}        & \multicolumn{1}{c|}{198}  & \multicolumn{1}{c|}{0.030} & \multicolumn{1}{c|}{0.130}   & \multicolumn{1}{c|}{0.020}  & \multicolumn{1}{c|}{0.028}  & \multicolumn{1}{c|}{0.001}    & \multicolumn{1}{c|}{0.127}  & \multicolumn{1}{c|}{0.010} & \multicolumn{1}{c|}{0.497} &  &  \\ \cline{2-11}
    \multicolumn{1}{|c|}{}                              & \multicolumn{1}{c|}{$D_l$}         & \multicolumn{1}{c|}{198}  & \multicolumn{1}{c|}{0.01}  & \multicolumn{1}{c|}{234.69}  & \multicolumn{1}{c|}{26.46}  & \multicolumn{1}{c|}{39.88}  & \multicolumn{1}{c|}{1590.58}  & \multicolumn{1}{c|}{6.93}   & \multicolumn{1}{c|}{13.65} & \multicolumn{1}{c|}{2.47}  &  &  \\ \cline{1-11}                                                     
    \end{tabular}}
    \label{T:SSMD}
\end{table}

\subsection{Evaluation on previous studies}\label{sec:evaluation}
Although extensive research has been carried out on $D_l$, there is limited awareness of the need for objective and
comprehensive comparison of these works. This makes the understanding of connections among those research results unclear. 
For example, it is hard to judge
which type of method is more suitable for the prediction of $D_l$ based on separate studies. 
The main obstacle is the lack of an appropriate, adequate database. 
However, datasets used in previous research are not the same. 
Most of the datasets are small, only 
dozens or hundreds of samples, which lack representativeness. The missing of proper data cleaning can easily make the previous evaluation biased. 
Defects in the database make the simple comparison of evaluation indexes in different studies meaningless. 
The uncertainty in model training also creates difficulties. In application of those methods, the selection of hyperparameters is often done by 
test and trail. This could cause results obtained with the same method distinctly, which will make it very hard to carry out proper performance comparison. 

To carry out a reliable comparison, these two problems need to be solved. The first problem can be easily 
tackled with the development of datasets in the previous two sections. The dataset in this paper is fairly large and cleaned 
carefully. It can fully reflect the potential performance of models in real life. 
As for the second problem, it will be solved through a general assessment. SVM, ANFIS and NN are the most commonly used 
approaches in implicit ML-driven studies. The main differences between similar studies in 
each approach are the hyperparameters of the model and used datasets. In our evaluation, the average performance 
is more concerned than the specific performance limits. Those different methods will be each implemented multiple times 
with random hyperparameters to get an average performance score. 
Then average scores of those models will be recorded to display their prediction abilities. 

The objectives in this evaluation are: statistical prediction formulas (9 equations, Table \ref{T:statisticalEq}), explicit ML-driven prediction formulas (9 equations, 
Table \ref{T:explicitEq}) and implicit ML-driven prediction models (SVM, ANFIS, NN, and ElasticNet). An internal evaluation of each method 
will be first made. Then the best two models in each method will be selected for further external comparison. 

The evaluation metrics used are Root Mean Squared Error (RMSE), Mean Absolute Percentage Error (WMAPE), 
R-Square ($R^2$) and Discrepancy Ratio (DR). See definitions of these metrics 
in supplementary material, Section S3. 

\begin{figure}[]
    \begin{centering}
    \includegraphics[width=0.47\linewidth]{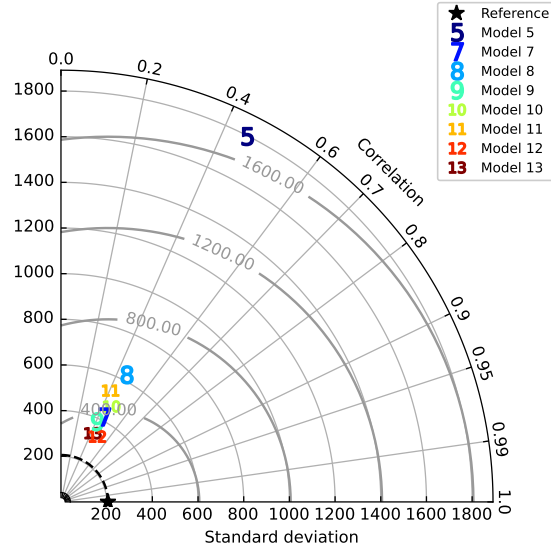}
    \caption{The Taylor diagram of statistical models} 
    \label{fig:statisticalEva}
    Model 5 - Fischer / 1975; Model 7 - Liu et al. / 1977; Model 8 - Seo and Cheong / 1998; Model 9 - 
    Koussis and Rodr´ıguez-Mirasol / 1998; Model 10 - Deng et al. / 2001; Model 11 - Kashefipour and 
    Falconer / 2002; Model 12 Zeng and Huai / 2014; Model 13 - Disley et al. / 2014.
    \end{centering}    
\end{figure}

The evaluation result of statistical models is illustrated in Table \ref{T:statisticalEva} and the Taylor diagram 
in Fig. \ref{fig:statisticalEva} (Model 
6 is not included due to the paucity of data on the energy slope). 
The result shows that RMSE values are quite high.
Most WMAPE values are over 1.00 and $R^2$ negative. The average accuracy of prediction is lower than 40\%. The best two models 
in the statistical method are Zeng and Huai / 2014 (Model 8 
in Fig. \ref{fig:statisticalEva}) and Disley et al. / 2014 (Model 9 in Fig. \ref{fig:statisticalEva}). All these metrics signal 
that the extrapolation ability of statistical models is poor.
This is mainly because the complexity of the $D_l$ problem has already surpassed the regression-ability limit of the statistical method.

\begin{table}[]
    \caption{The evaluation of statistical models}
    \resizebox{\textwidth}{!}{
    \begin{tabular}{|c|c|c|c|c|c|c|c|c|}
    \hline
    \multirow{2}{*}{Seq} & \multirow{2}{*}{RMSE} & \multirow{2}{*}{WMAPE} & \multirow{2}{*}{R2} & \multicolumn{4}{c|}{DR}                                                                                                & \multirow{2}{*}{\begin{tabular}[c]{@{}c@{}}Accuracy\\ (-0.3\textless{}dr\textless{}0.3)\end{tabular}} \\ \cline{5-8}
                      &                       &                    &                     & dr $\leq$ -0.3              & -0.3 < dr $\leq$ 0            & 0 < dr $\leq$ 0.3              & dr > 0.3 &                                                   \\ \hline
    Model 5           & 1758.58               & 4.81               & -72.28              & 0.24                        & 0.13                          & 0.18                         & 0.45                      & 30.91                              \\ \hline
    Model 7           & 378.45                & 1.40               & -2.39               & 0.18                        & 0.18                          & 0.28                         & 0.35                      & 46.82                              \\ \hline
    Model 8           & 572.65                & 1.57               & -6.77               & 0.13                        & 0.14                          & 0.28                         & 0.45                      & 42.12                              \\ \hline
    Model 9           & 362.96                & 1.49               & -2.12               & 0.12                        & 0.15                          & 0.30                         & 0.43                      & 44.70                              \\ \hline
    Model 10           & 423.51                & 1.23               & -3.25               & 0.15                        & 0.16                          & 0.27                         & 0.43                      & 42.42                              \\ \hline
    Model 11           & 491.58                & 1.29               & -4.73               & 0.22                        & 0.28                          & 0.16                         & 0.34                      & 43.79                              \\ \hline
    Model 12           & 291.60                & 0.96               & -1.01               & 0.19                        & 0.22                          & 0.25                         & 0.35                      & 46.52                              \\ \hline
    Model 13           & 309.66                & 1.01               & -1.27               & 0.12                        & 0.25                          & 0.24                         & 0.38                      & 49.55                              \\ \hline
    \end{tabular}
    }
    \label{T:statisticalEva}
\end{table}

\begin{figure}[]
    \begin{centering}
    \includegraphics[width=0.5\linewidth]{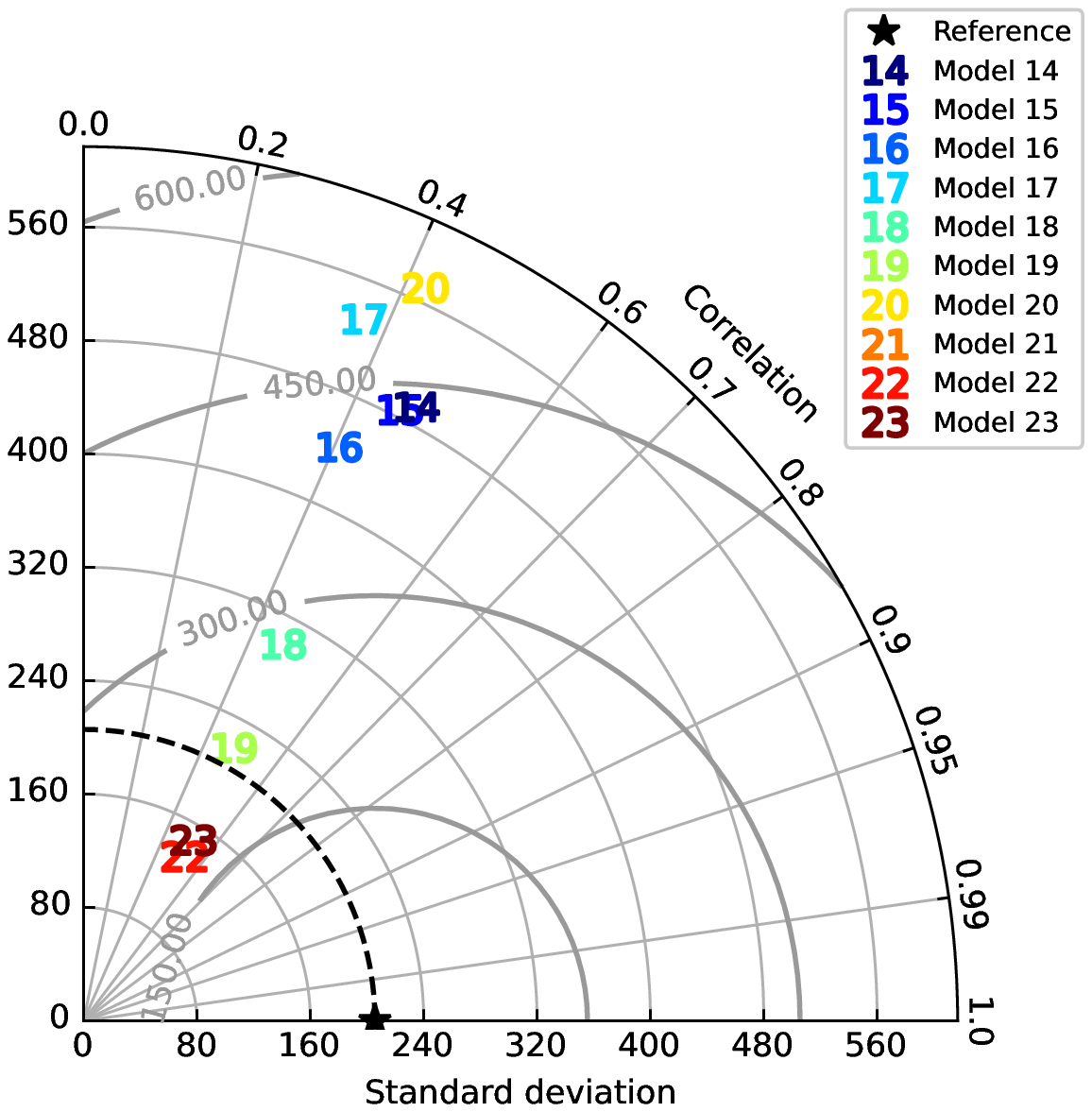}
    \caption{The Taylor diagram of the explicit ML-driven models}
    where Model 14 - Sahay and Dutta / 2009; Model 15 - Li et al. / 2013; Model 16, 17 - Scatter and Gharabaghi / 2015; 
    Model 18 - Wang and Huai / 2016; Model 19 - Wang and Huai / 2017;
    Model 20 - Alizadeh et al. / 2017; Model 21 - Riahi-Madvar / 2019; Model 22, 23 - Memarzadeh R., et al. / 2020; 
    Model 24, 25 - Riahi-Madvar et al. / 2020. 
    \label{fig:explicitMLEva}
    \end{centering}    
\end{figure}

Compared with statistical methods, explicit ML-driven methods achieve much better results. Due to stronger regression ability, 
all indexes of those studies are significantly improved (Table \ref{T:explicitMLEva} and Fig. \ref{fig:explicitMLEva}). 
The best model (Memarzadeh R., et al. / 2020, model 9 in Fig. \ref{fig:explicitMLEva}) 
achieves R2 of 0.22 and prediction accuracy of 55\%. 
The RMSE and WMAPE values decrease to 181.02 and 0.72. This shows that the prediction deviation is in a reasonable range. 
The assessment result shows that the explicit ML-driven method has better generalization ability than the statistical method. It can 
learn more patterns from samples 
and function more precisely in extrapolation prediction. The best two models in this branch are model 9 and 10 of 
Memarzadeh R., et al. / 2020. It is worth noting that the results of model 11 and model 12 are very poor. 
After verification, they are developed by combining the optimization algorithm and NN on a small data set (71 
samples from \cite{tayfur2005predicting}). 
$R^2$ values in paper of Riahi-madvar\cite{madvar2020derivation} are 0.94 and 0.81 for training and testing sets, which are very high. 
This poor assessment result reveals that these two models suffer from critical over-fitting 
problem\cite{bishop2006pattern, haykin1998neural}. 
The prediction result from these two models is biased towards the training dataset, which makes them lack applicable values.
Therefore, model 11 and model 12 are not included in the visualization.

\begin{table}[]
    \caption{The evaluation of explicit ML-driven models}
    \resizebox{\textwidth}{!}{
    \begin{tabular}{|c|c|c|c|c|c|c|c|c|}
    \hline
    \multirow{2}{*}{Seq} & \multirow{2}{*}{RMSE} & \multirow{2}{*}{WMAPE} & \multirow{2}{*}{R2} & \multicolumn{4}{c|}{DR}                                                                                                & \multirow{2}{*}{\begin{tabular}[c]{@{}c@{}}Accuracy\\ (-0.3\textless{}dr\textless{}0.3)\end{tabular}} \\ \cline{5-8}
                      &                       &                      &                     & dr $\leq$ -0.3              & -0.3 < dr $\leq$ 0            & 0 < dr $\leq$ 0.3              & dr > 0.3 &                                                   \\ \hline
    Model 14           & 439.56                & 1.28              & -3.58               & 0.15                        & 0.19                          & 0.23                         & 0.43                      & 41.97                                                                                                 \\ \hline
    Model 15           & 434.98                & 1.21               & -3.48               & 0.19                        & 0.20                          & 0.24                         & 0.37                      & 44.24                                                                                                 \\ \hline
    Model 16           & 408.89                & 1.18               & -2.96               & 0.13                        & 0.22                          & 0.28                         & 0.37                      & 49.85                                                                                                 \\ \hline
    Model 17           & 499.63                & 1.32              & -4.92               & 0.09                        & 0.26                          & 0.25                         & 0.41                      & 50.30                                                                                                 \\ \hline
    Model 18           & 274.09                & 0.94               & -0.78               & 0.19                        & 0.21                          & 0.26                         & 0.33                      & 47.27                                                                                                 \\ \hline
    Model 19           & 217.01                & 0.84                & -0.12               & 0.21                        & 0.25                          & 0.22                         & 0.32                      & 46.97                                                                                                 \\ \hline
    Model 20           & 521.13                & 1.25               & -5.44               & 0.23                        & 0.26                          & 0.26                         & 0.25                      & 51.82                                                                                                 \\ \hline
    Model 21           & 1429.85               & 3.31              & -47.45              & 0.17                        & 0.20                          & 0.24                         & 0.39                      & 43.48                                                                                                 \\ \hline
    Model 22           & 181.02                & 0.72                & 0.22                & 0.27                        & 0.30                          & 0.25                         & 0.18                      & 55.45                                                                                                 \\ \hline
    Model 23          & 183.13                & 0.74                & 0.21                & 0.22                        & 0.28                          & 0.21                         & 0.29                      & 49.09                                                                                                 \\ \hline
    Model 24          & 1010.22               & 10.35               & -23.18              & 0.07                        & 0.01                          & 0.02                         & 0.91                      & 2.73                                                                                                  \\ \hline
    Model 25          & 906.64                & 9.45               & -18.48              & 0.03                        & 0.03                          & 0.03                         & 0.91                      & 6.21                                                                                                  \\ \hline
    \end{tabular}
    }
    \label{T:explicitMLEva}
\end{table}

\begin{figure}[]
    \begin{centering}
    \includegraphics[width=0.5\linewidth]{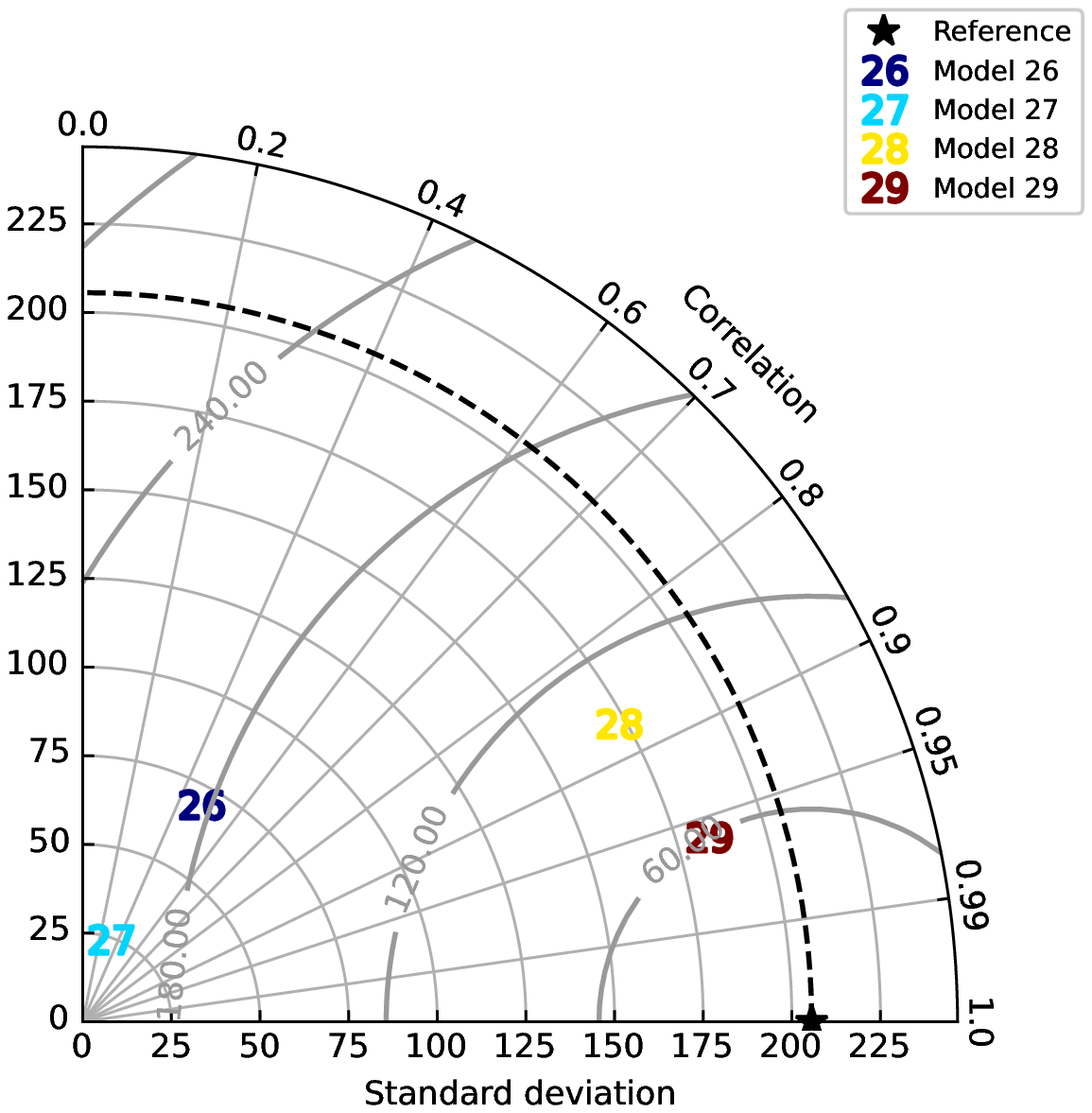}
    \caption{The Taylor diagram of the implicit ML-driven models}
    where Model 26 - NN; Model 27 - ANFIS; Model 28 - ElasticNet; Model 29 - SVM. 
    \label{fig:implicitMLEva}
    \end{centering}    
\end{figure}

The result of implicit ML-driven methods is very satisfying(Table \ref{T:implicitMLEva} and Fig. \ref{fig:implicitMLEva}). 
The best model is the support vector machine (SVM, model 29 in Fig. \ref{fig:implicitMLEva}). The $R^2$ is about 0.36 
and prediction accuracy over 60\%, which indicate
SVM has very high practical worth. However, NN and ANFIS, which should have stronger regression ability, underperform than SVM and ElasticNEt. 
After analysis, this is mainly due to 
two reasons. Firstly,  Network design of NN and ANFIS has very strong randomness\cite{gakuto2019method}. There are no rules or modes for development 
at present and hyperparameters are usually selected by test and trial. This makes the model performance fluctuate in a certain range. 
It is possible to find a better model through costly hyperparameter selection. But as mentioned before, the average performance is the core of this section. 
A well-tuned NN or ANFIS model can only 
show that model is better but not the method. Secondly, stronger 
regression ability does not mean a superior model.
Although detailed data cleaning is carried out, there are still some noises in samples. It can be inferred that NN tried to learn from those noises. 
In small dataset, NN can learn those noises very well\cite{madvar2020derivation} and then be prone to overfitting. 
Quite satisfying results can be obtained. However, samples from different experiments will have different error distributions. When they come together, 
NN will try to learn all these errors, which causes the final result to be unsatisfying.

\begin{table}[]
    \caption{The evaluation of implicit ML-driven models}
    \resizebox{\textwidth}{!}{
    \begin{tabular}{|c|c|c|c|c|c|c|c|c|}
    \hline
    \multirow{2}{*}{Seq} & \multirow{2}{*}{RMSE} & \multirow{2}{*}{WMAPE} & \multirow{2}{*}{R2} & \multicolumn{4}{c|}{DR}                                                                                                & \multirow{2}{*}{\begin{tabular}[c]{@{}c@{}}Accuracy\\ (-0.3\textless{}dr\textless{}0.3)\end{tabular}} \\ \cline{5-8}
                      &                       &                      &                     & dr \textbackslash{}leq -0.3 & -0.3\textless{}dr\textless{}0 & 0\textless{}dr\textless{}0.3 & dr \textbackslash{}beq0.3 &                                                                                                       \\ \hline
    Model 1           & 180.18                & 0.79                & 0.10               & 0.10                        & 0.32                          & 0.20                         & 0.38                      & 52.02                                                                                                 \\ \hline
    Model 2           & 199.30                & 0.76                & 0.13                & 0.16                        & 0.10                          & 0.20                         & 0.54                      & 53.55                                                                                                 \\ \hline
    Model 3           & 101.27                & 0.42                & 0.47                & 0.21                        & 0.38                          & 0.22                         & 0.20                      & 59.60                                                                                                 \\ \hline
    Model 4           & 59.50                 & 0.32                & 0.51                & 0.14                        & 0.34                          & 0.26                         & 0.26                      & 68.80                                                                                                 \\ \hline
    \end{tabular}}
    \label{T:implicitMLEva}
\end{table}

\begin{figure}[]
    \begin{centering}
    \includegraphics[width=0.5\linewidth]{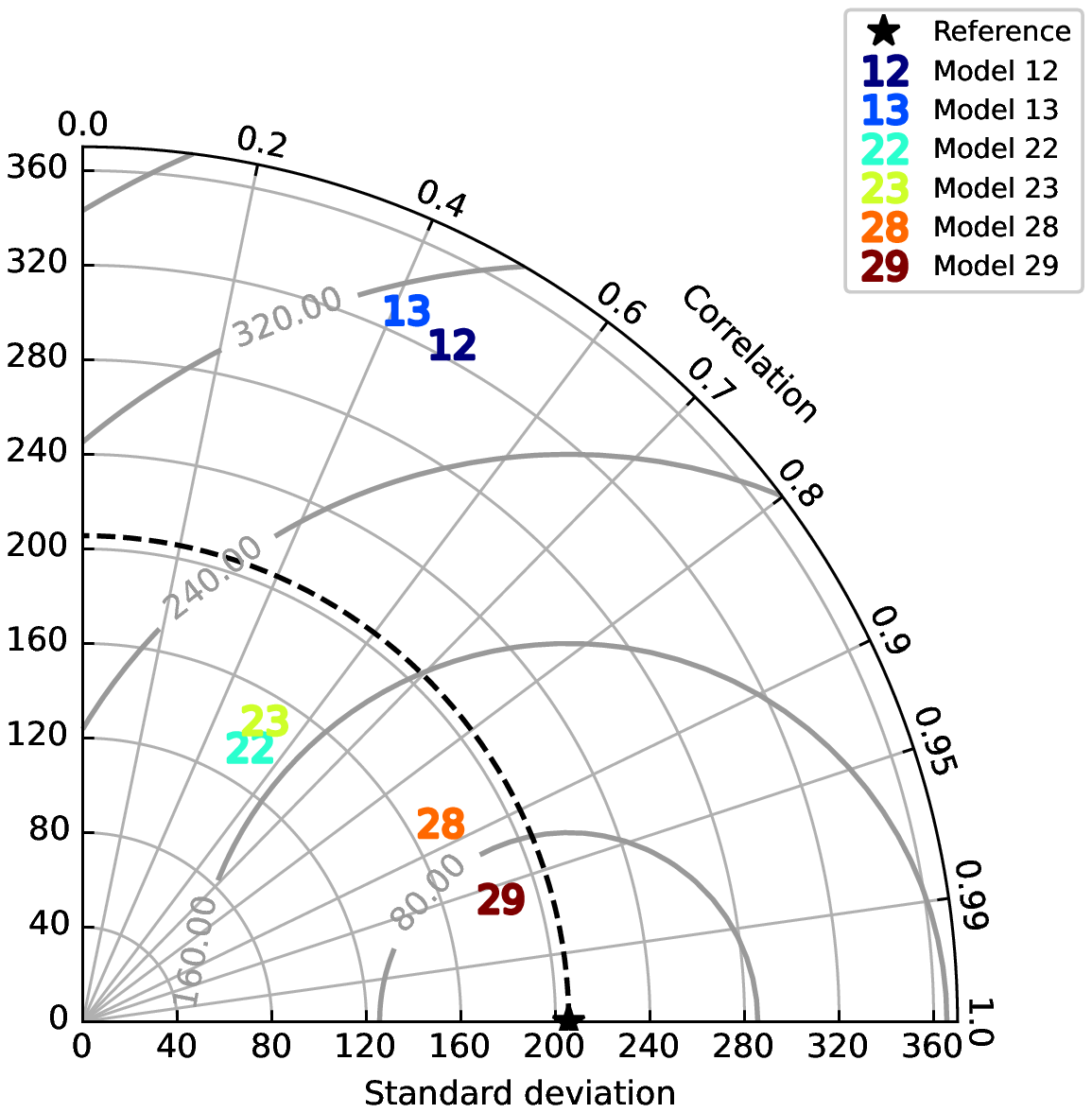}
    \caption{The Taylor diagram of the state-of-art models in each branches}
    where Model 12 - Zeng and Huai / 2014; Model 13 - Disley et al. / 2014; Model 22, 23 - Memarzadeh R., et al.; Model 28 - ElasticNet; Model 29 - SVM.
    \label{fig:selectedEva}
    \end{centering}    
\end{figure}

Through visualizing the best two models of each method (Fig. \ref{fig:selectedEva}), the rank of three regression techniques is: 
Implicit ML-driven methods $>$ Explicit ML-driven methods $>$ 
statistical methods. Among them, statistical methods have limited extrapolation ability and implicit 
ML-driven methods suffer from their black-box nature,
which make the Explicit ML-driven method to
be the most proper solution to gain a fair compromise between predictability and understanding on this phenomenon.
Optimization algorithms are popular methods in this field. But the regression problem is not a good topic for 
them\cite{goldberg1989genetic, boyd2004convex, nocedal2006numerical}.
Besides, there is still some performance distance between explicit ML-driven methods and 
implicit ML-driven methods (Fig. \ref{fig:selectedEva}), which indicates room for 
improvement in explicit ML-driven methods.

\begin{figure}[]
    \begin{centering}
    \includegraphics[width=0.75\linewidth]{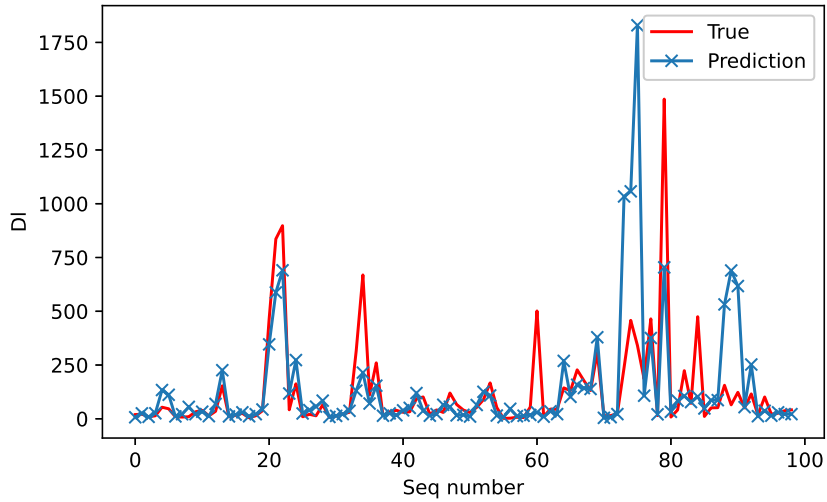}
    \caption{Point to point plot of randomly selected samples}
    \label{fig:p2pPlot}
    \end{centering}    
\end{figure}

It is interesting to note that the evaluation result of those models is much lower than the value reported in their papers. 
A common reason for all cases is the size effect 
of the training dataset. By plotting 100 prediction values point to point 
randomly (Fig. \ref{fig:p2pPlot}), it can be found that the main error sources 
are some extreme predictions. 
In the development of those models, the lack of outlier-cleaning and subset selection is common. 
Besides, the used datasets are usually small, only dozens of hundreds of samples, which 
can easily lead to a biased result. 
In the general form of $D_l$ models (Eq. \ref{eq:generalForm}), $d$ and $U^*$ are divisible items (They appear in fraction denominators explicitly). 
With many small values in $d$ and $U^*$, those biased models can generate huge predictive 
errors due to numberical singularities and deteriorate the model performance significantly. 

However, this does not mean a larger dataset is inappropriate. A large dataset can bring advantages in both establishment and evaluation 
of the model. 
A larger dataset can also enable the model to discover more and better patterns, which guarantees the model to have a proper 
generalization ability. Taking models in \cite{memarzadeh2020a} as examples, an explicit prediction model is developed
on 503 samples and this model shows much better prediction and generalization ability 
than all other symbolic models. Apart from its advantages in the algorithm, a strong data basis is the 
key to its performance advantage. 
Moreover, The application range of $D_l$ prediction models will usually exceed the training dataset. It is difficult to draw 
effective evaluation conclusions from small datasets. On the contrary, a large dataset can give a more 
comprehensive result of 
the model under different circumstances, especially the biased situation. 
In conclusion, a large and convincing dataset is needed for the regression of the $D_l$. 

Another factor concerns the characteristics of regression methods themselves. Depending on the regression ability, 
those regression methods can be split into three 
levels. The statistical method has weak regression ability and cannot learn many patterns from the data, which leads 
to poor generalization ability. On the contrary, 
some implicit ML-driven methods, such as NN have very strong regression ability. However, this advantage didn't bring 
much improvement in performance. According to the 
low RMSE value but small R2 of NN in evaluation result compared with model 9 in Table \ref{T:explicitMLEva}, it 
can be inferred that these methods attempt to learn from the noise instead of physical patterns. In addition, there are 
not many empirical paradigms on the training of NN and other similar network models, which makes overfitting a common 
phenomenon. This can lead to problems with good 
training results but poor testing results. Interestingly, methods with medium regression ability achieve much better results, 
such as explicit ML-driven methods and 
simplified implicit ML-driven methods (SVM and ElasticNet). The regression ability of these methods is 
better than that of statistical methods. But they do not reach 
the level of depicting the noise, which brings an excellent generalization and extrapolation ability.

\section{Methods}\label{sec:method}

Evolutionary symbolic regression network (ESRN) proposed in this paper is a novel explicit ML-driven method, 
which combines the genetic algorithm (GA) and the neural network. The detialed information about these two 
algorithms are included in the supplementary material, Section S4. 

The main idea of this algorithm is to search for the global optimal solution with GA on the basis of local 
optimal solutions provided by NN. That is because gradient optimization of NN will have its 
defects on discrete problems, while GA can bridge this problem with its 
applicability in both discrete and continuous situations. The origin of this idea is from two characteristics of NN: 

I. NN can represent many kinds of formulas

\begin{figure}[]
    \begin{centering}
    \includegraphics[width=0.62\linewidth]{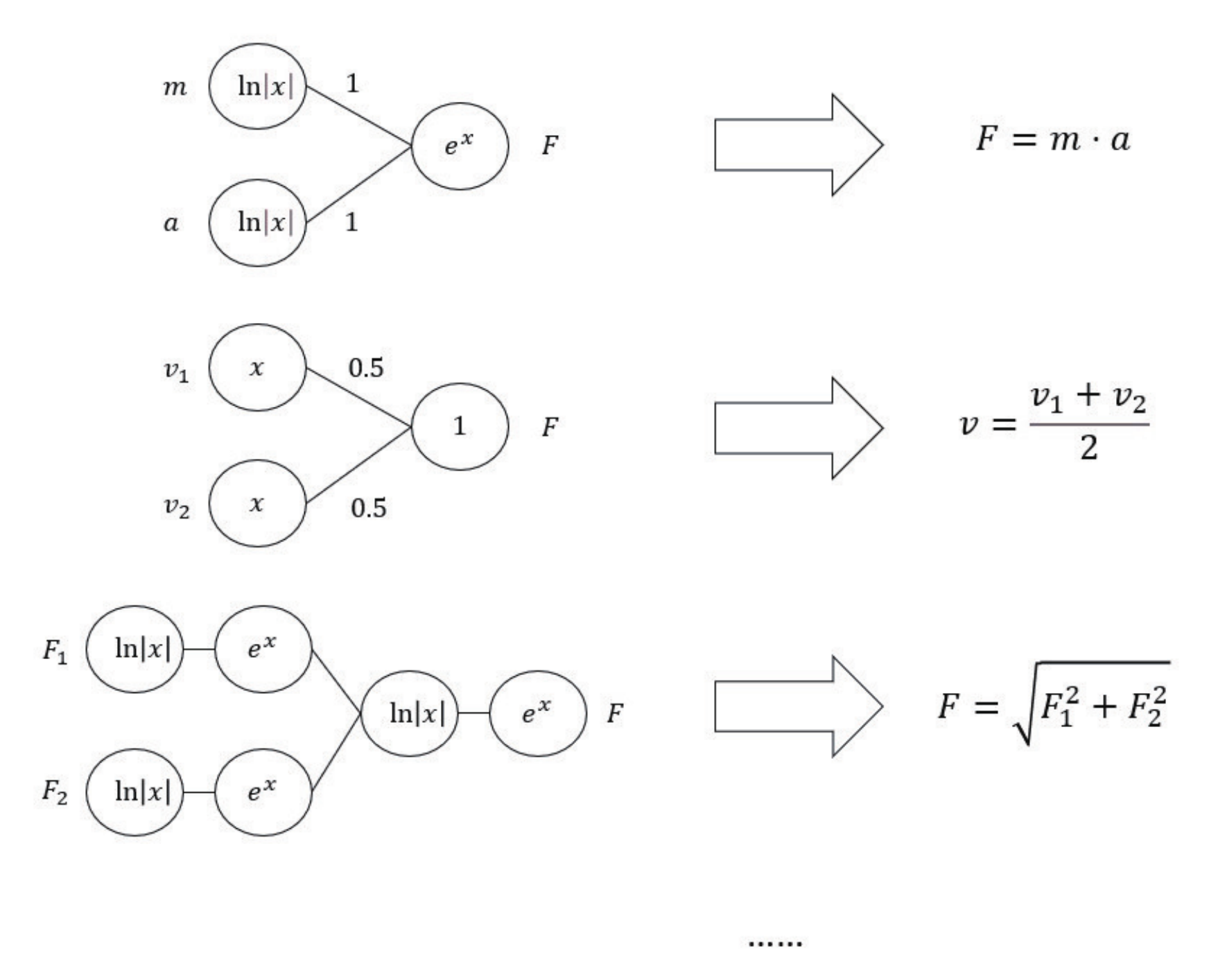}
    \caption{Examples of expressing formulas with NN}
    \label{fig:NNrep}
    \end{centering}    
\end{figure}

With a well defined activation table (Table \ref{T:actiLt}), all kinds of equations can be encapsulated 
into a network (Fig. \ref{fig:NNrep}). The topology and weights of NN will represent the combination and 
parameters of the equation separately. 
This makes NN an interpretable symbolic combination and easier to understand. 

\begin{table}[]
    \centering
    \setlength{\abovecaptionskip}{0pt}%
    \setlength{\belowcaptionskip}{10pt}%
    \caption{The list of selected activation function}
    \begin{tabular}{l}
    $a_{1}(x)=1$    \\
    $a_{2}(x)=x$    \\
    $a_{3}(x)=e^{x}$    \\
    $a_{4}(x)=ln(|x|)$   \\
    $a_{5}(x)=\frac{1}{1+e^{-x}}$   \\ 
    \end{tabular}
    \label{T:actiLt}
\end{table}

II. A strong regression ability

Universal Approximation Theorem\cite{hornik1989multilayer} reveals a striking feature of NN: for any distribution, 
there is always a corresponding network which can learn and 
imitate its patterns. This makes NN a robust solution for prediction.

\begin{figure}[]
    \begin{centering}
    \includegraphics[width=1\linewidth]{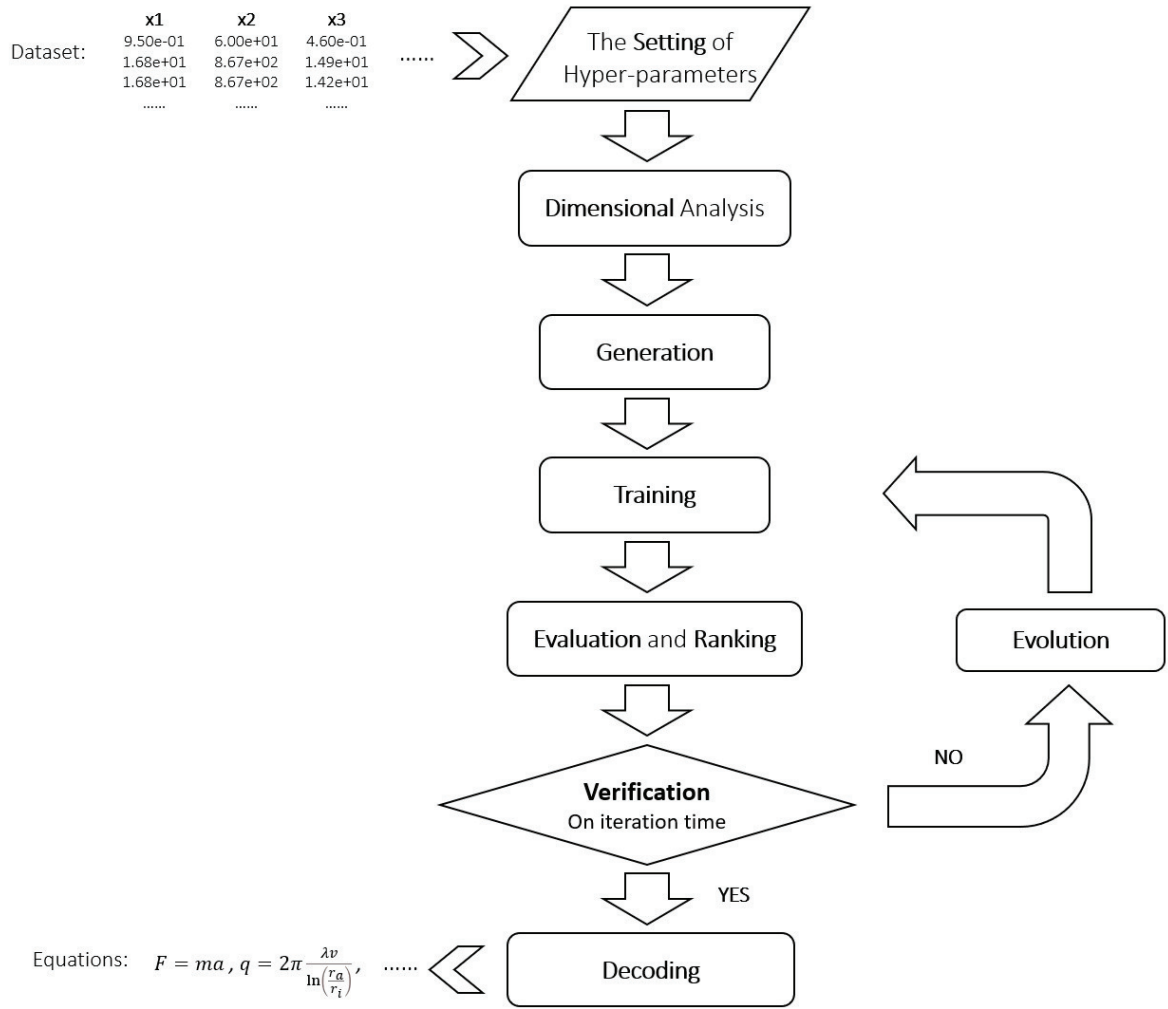}
    \caption{Flow chart of the evolutionary symbolic regression network}
    \label{fig:flowchart}
    \end{centering}    
\end{figure}

The general framework of this evolutionary regression network is listed as follows. For better understanding, 
A flow chart is provided (Fig. \ref{fig:flowchart}). 

(1) The setting of hyper-parameters: $N$ - the population size; $M$ - the evaluation metric; $T$ - the iteration time; $S_{n}$ - the topology restriction 
vector, which limits the number of neurons in each layer. (for example, [3,2,1] stands for a three-layer network, with 3 input neurons, 2 hidden 
neurons and 1 output neuron in each layer at most)

(2) Dimensional analysis(DA): Parameters for the $D_l$ prediction are $w(m)$, $d(m)$, $U(m/s)$, $U^*(m/s)$, which have known units.
Combined with the unit of those 
physical quantities, dimensional analysis is carried out to find possible parameter combinations.

(3) Generation: According to those combinations, $N$ neural networks with topology according to $S_{n}$ will be generated as initial population with random inputs and outputs. 

(4) Training: These candidates will be trained in batch with Adam. 

(5) Evaluation and ranking: Those NNs will be evaluated and ranked according to $M$. The first $N$ networks will be kept and the rest dumped. 

(6) Verification: the iteration time will be checked whether it is lower than $T$. If lower, go to 7; otherwise, go to 8. 

(7) Evolution: NNs generated in (3) will serve as the genome for GA evolution. 
The main evolution process refers to the topology modification of NNs according to mutation and crossover strategies. 
New candidates will be generated. The main purpose of this step is to expand the solution search space, 
which is the key to obtaining a global optimal solution. Then go to 3. 

(8) Decoding: The top NN will be transferred into a symbolic equation for further analysis.

To illustrate more details, processing on Dimensional Analysis, crossover, mutation and avoidance of overfitting
will be explained and discussed.

\textbf{Dimensional Analysis}: All physical phenomena can be described by governing equations, which always keep a same dimension 
at both sides of the formula. This property is called dimensional homogeneity. The homogeneity in dimensions often simplifies 
the problem and brings advantages to regression. Therefore, parameters from this problem 
are transferred into a dimensionless combination according to the Buckingham $\pi$ theorem. 
The $\pi$ theorem states that a function of n variables with m dimensional units (Eq. \ref{Eq: pi1}) can be reorganized as a similar 
one with (n - m) dimensionless 
varialbes (Eq. \ref{Eq: pi2}).

\begin{equation}
    y=f(x_1, x_2, ..., x_n), \text{with m dimensional units} 
    \label{Eq: pi1}
\end{equation}
where $x_i$ = the dimensional input; $y$ = the dimensional output

\begin{equation}
    \pi_y=f(\pi_1, \pi_2, ..., \pi_{n-m})   \label{Eq: pi2}
\end{equation}
where $\pi_y$ = the dimensionless output; $\pi_x$ = the dimensionless input

Unit matrix $M^u_x$ is defined as a combination of 
unit vector $u_i$ corresponding to the input variable $x_i$ and unit matrix $M^u_y$ corresponding to the output variable $y_i$. 
$N$ and $p$ are solutions to equation $M^u_xN=0$ and $M^u_xp=M^u_y$. 
Then Eq. \ref{Eq: pi1} and Eq. \ref{Eq: pi2} can be connected with the relationship stated in Eq. \ref{Eq: pi3}.  

\begin{gather}
    \pi_{xi} = \prod_{j=1}^{n} x_{j}^{N^{ij}},   \notag\\
    \pi_y=\frac{y}{\pi_y^*}, \quad \text{where} \; \pi_y^* = \prod_{j=1}^{n} x_{j}^{p^{j}}           
\label{Eq: pi3}
\end{gather}

The physical units of variables in this prediction problem are listed in Table \ref{T:unitTable}. Since there are multiple variables with the same unit, 
the possible combination can be enlarged by replacing $p$ with a linear combination of $p$ and $N$. Through above operations, the parameter 
candidates for further analysis are: $\frac{w}{d}$, $\frac{U}{U^*}$ and constant 1 for input; $\frac{D_l}{wU}$, 
$\frac{D_l}{wU^*}$, $\frac{D_l}{dU}$, $\frac{D_l}{dU^*}$ for output. 

\begin{table}[]
    \centering
    \caption{Unit table of parameters}
    \begin{tabular}{|c|c|c|c|}
    \hline
    Parameter & Unit & L & T \\\hline
    $w$, $d$      & m    & 1 & 0  \\\hline
    $U$, $U^*$    & m/s  & 1 & -1  \\\hline
    \end{tabular}\label{T:unitTable}
\end{table}

The crossover and mutation are the key for this algorithm to evolve expression and find the global optimal solution. 
Randomness will be utilized in the following parts frequently.  It can guarantee a exploration of existing form (\ref{eq:generalForm}) 
and other possible combinations in the solution space. The introduce of randomness is designed to 
bring diversity into the search of expression. Therefore, the specific form of randomness is not important, 
as long as it can create difference. In this framework, random values with uniform distribution are 
selected and used in all cases.

\begin{figure}[]
    \begin{centering}
    \includegraphics[width=0.8\linewidth]{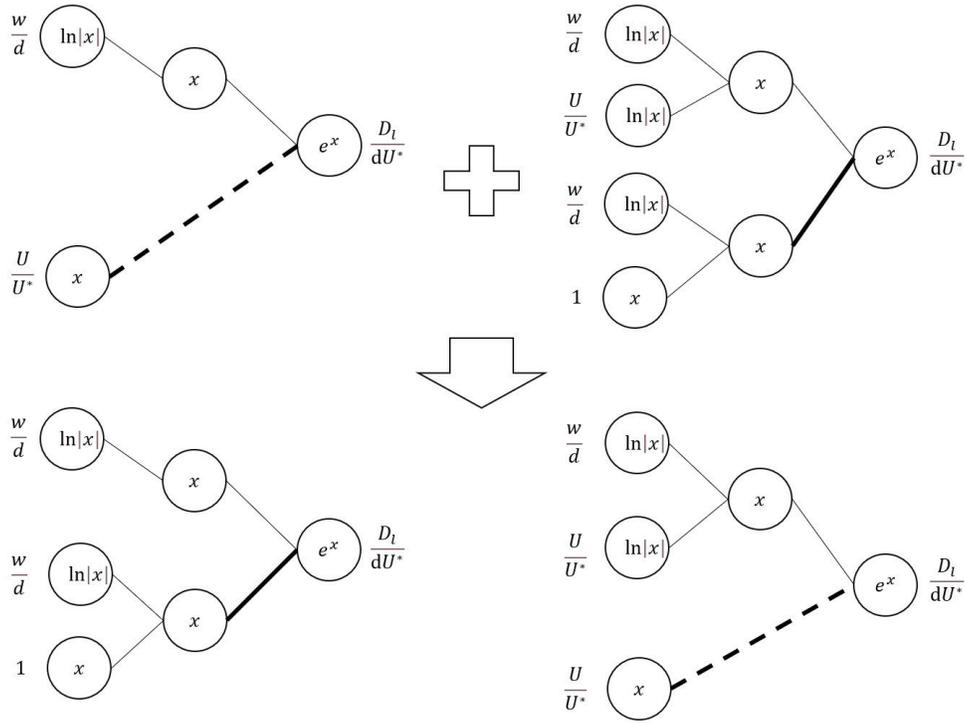}
    \caption{The crossover strategy}
    \label{fig:crossover}
    \end{centering}    
\end{figure}

\textbf{Crossover}: The crossover strategy focuses on finding a better topology combination in the existing group. 
In ESRN, sub-networks connected to the ouput neuron
will be exchanged randomly between different candidates NNs (Fig. \ref{fig:crossover}). Through generations of selection, better  
topologies will be obtained and worse discarded.  

\textbf{Mutation}:
Mutation is a more radical strategy. It can enlarge the solution search space and bring unexpected results. 
Two modes are utilized here, which are activation mutation
and and candidate mutation. 

\begin{figure}[]
    \begin{centering}
    \includegraphics[width=1\linewidth]{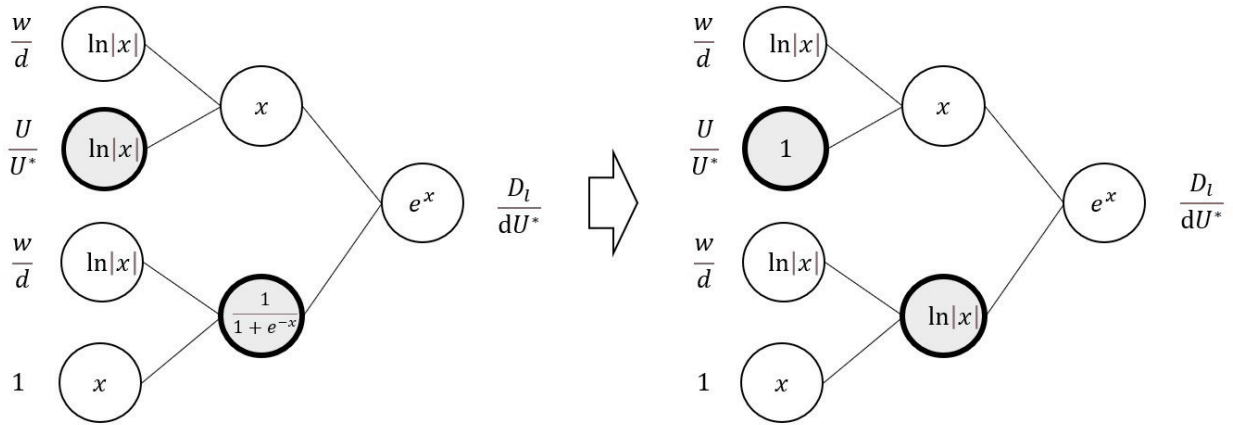}
    \caption{The activation mutation}
    \label{fig:mutation1}
    \end{centering}    
\end{figure}

Activation mutation (Fig. \ref{fig:mutation1}) refers to stochastic modification of the activation function in neurons. It can bring 
a new combination of equations into the system. 
All neurons in the network are able to have this mutation. Both original 
and mutated networks will be kept in the population. 

\begin{figure}[]
    \begin{centering}
    \includegraphics[width=1\linewidth]{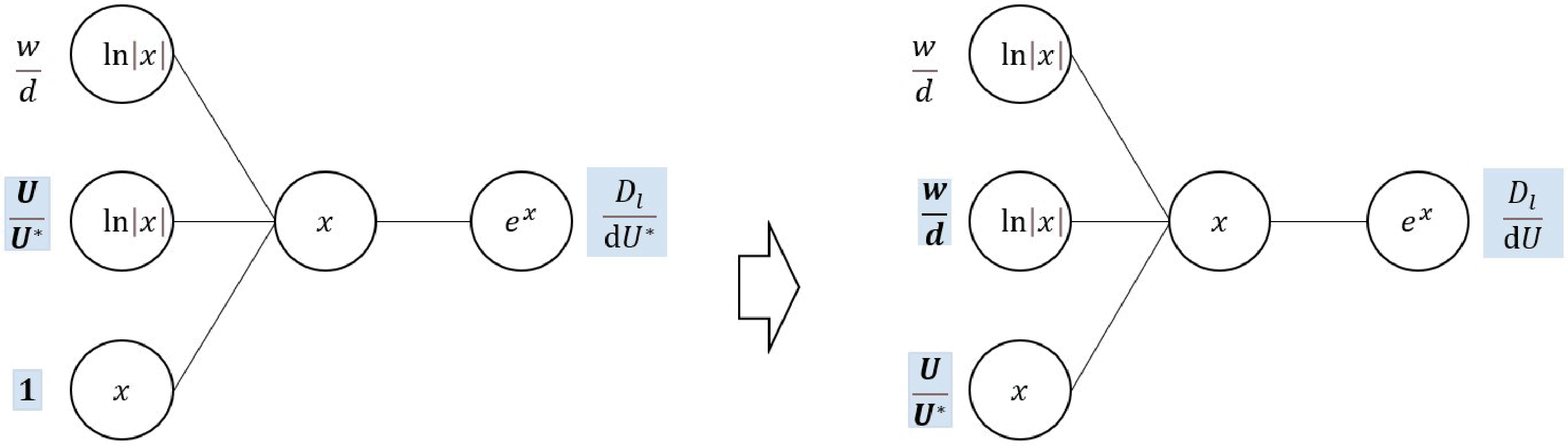}
    \caption{The candidate mutation}
    \label{fig:mutation2}
    \end{centering}    
\end{figure}

Candidate mutation (Fig. \ref{fig:mutation2}) refers to a random change of input and output dimensionless candidates mentioned 
in dimensional analysis. It will adjust those dimensionless combinations randomly 
within each branches and enable the algorithm to explore more input-output options. 
The processing strategy is the same as the activation mutation. Both origin and the mutation will both be kept.

\textbf{Avoidance of overfitting}:
As mentioned, overfitting is a frequent problem in existing
studies. It can degrade the generalization ability of the model severely.
To avoid overfitting, it is crucial to acquire the performance of a model on both training and testing sets.
Therefore, a $R^2$-generation plot will be proposed. This plot is a visualization
of the top model's $R^2$ in each generation and the generation time during training and testing.
It can reveal the trend of the model's generalization performance. Through artificial selection based on the trend,    
the optimal model can be obtained.
The reason for adopting $R^2$ as the evaluated metric is the common existence of extreme predictions (Fig. \ref{fig:p2pPlot}).
Other metrics such as MSE are sensitive to those extreme values and cannot fully
reflect the performance of the model. On the contrary, $R^2$ is a normalized measure, which
can consider the difference not only between true and predictive
values but also predictions and their average. It is less affected by those extreme values.

\section{Results and evaluations}\label{sec:result}
Hyper-parameters used in this experiment are listed in Table \ref{T:hyperTable}.

\begin{table}[]
    \centering
    \caption{The table of hyper-parameters}
    \begin{tabular}{|c|c|c|c|}
    \hline
    $N$    & $S_n$    & $T$   & $M$      \\\hline
    100    & [5,3,1]  & 200   & $R^2$  \\\hline
    \end{tabular}\label{T:hyperTable}
\end{table}

\begin{figure}[]
    \begin{centering}
    \includegraphics[width=0.75\linewidth]{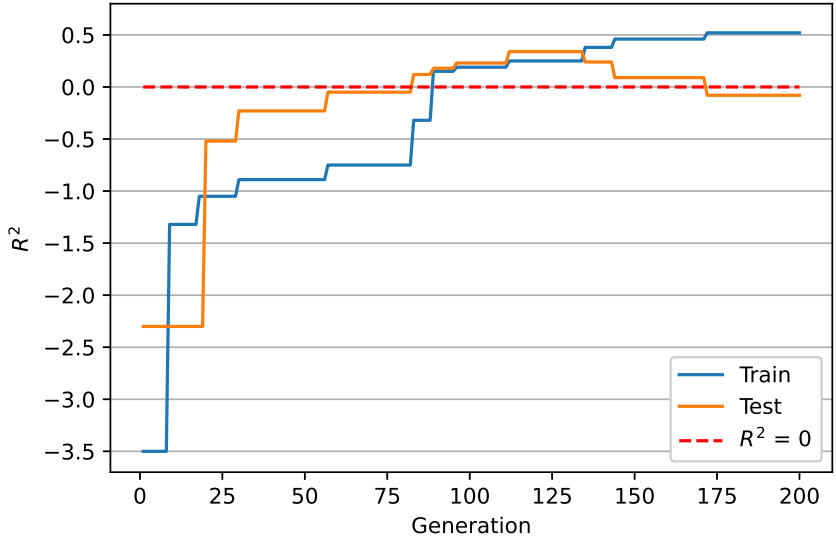}
    \caption{The $R^2$-generation plot during the implementation of ESRN}
    \label{fig:lossPlot}
    \end{centering}    
\end{figure}

The $R^2$-generation plot during the implementation of ESRN is shown in Fig. \ref{fig:lossPlot}. 
It is clear that $R^2$ of 
the training set keeps rising with the increase of generation, while $R^2$ of the testing set first increases and 
then decreases. It reaches the peak value from 112 to 135 generation. This reveals that top models after 135 generation are 
overfitting and the best model locates between 112 to 135 generation.   

\begin{figure}[]
    \begin{centering}
    \includegraphics[width=0.5\linewidth]{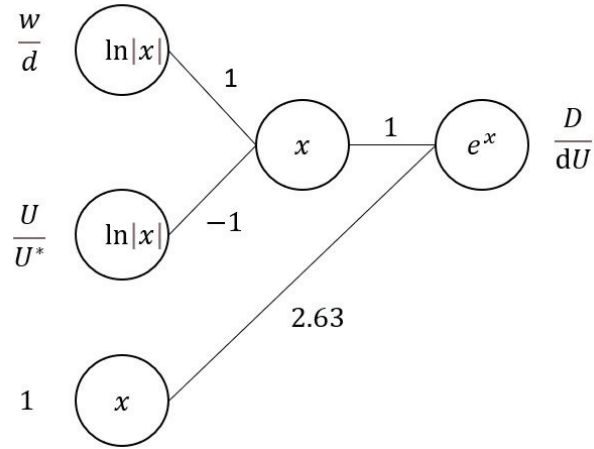}
    \caption{The topology of the result}
    \label{fig:resTopology}
    \end{centering}    
\end{figure}

The model distilled by ESRN during 112 to 135 generations is very elegant. The optimal network topology is shown 
in Fig. \ref{fig:resTopology}. Basically, it shares a similar topology with previous studies (Eq. \ref{eq:generalForm}). 
The corresponding equation converted from the topology is shown in Eq. \ref{Eq: res}.

\begin{gather}
    \frac{D_l}{dU}=e^{1 \times ln|\frac{w}{d}| - 1 \times ln|\frac{U}{U^*}|+1 \times 2.63 }\\   
    \frac{D_l}{dU}=e^{2.63} (\frac{w}{d})^{1} (\frac{U}{U^*})^{-1}\\
    D_l=13.89wU^*
    \label{Eq: res}
\end{gather}

To evaluate this model, performance metrics, scatter plot of observations and predictions, DR distribution plot and accuracy plot are calculated and implemented 
on the testing set.

The state of art research in explicit ML-driven method, 
two models of Memarzadeh R., et al. / 2020, are selected for comparison.

The performance metrics of these 3 models on the testing set are listed 
in Table \ref{T: pm}. All Metrics are in reasonable ranges. 
The errors of two models from \cite{RasoulMemarzadeh.2020} are almost the same, 
which is consistent with the previous comparison in Section \ref{sec:data}. 
The ESRN model shows superiorities in all indexes.

\begin{table}[]
    \centering
    \caption{Table of the performance metrics}
    \begin{tabular}{|c|c|c|c|}
    \hline
    Model      & RMSE  & WMAPE & $R^2$ \\ \hline
    ESRN           & 32.29 & 0.52  & 0.34                 \\ \hline
    $Model_{22}$   & 38.00 & 0.58  & 0.11                 \\ \hline
    $Model_{23}$   & 38.45 & 0.59  & 0.09                 \\ \hline
    \end{tabular}
    \label{T: pm}
\end{table}
Where $Model_{22}$ - Memarzadeh R., et al. / 2020; $Model_{23}$ - Memarzadeh R., et al. / 2020.

\begin{figure}[]
    \begin{centering}
    \includegraphics[width=1\linewidth]{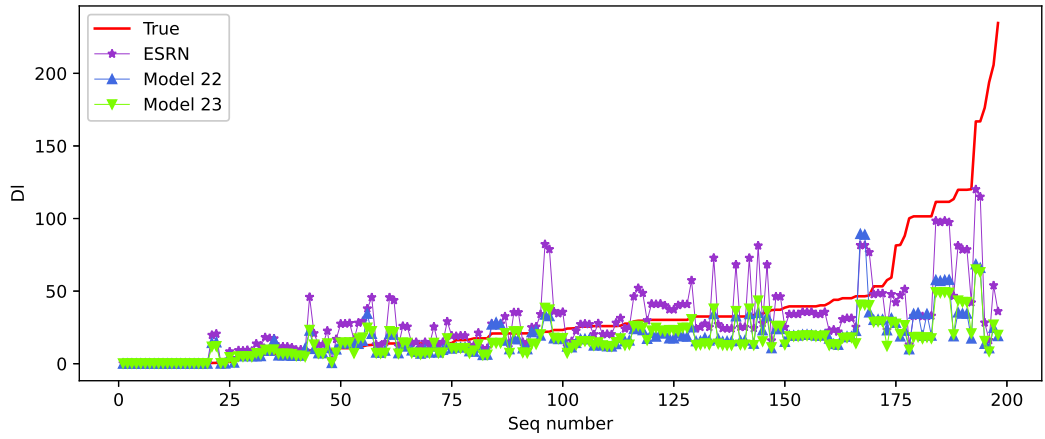}
    \caption{The scatter plot of true and predicted values}
    \label{fig:p2p}
    \end{centering}    
\end{figure}

Fig. \ref{fig:p2p} is the scatter plot of true and predicted values from different models. To make the result more intuitive, 
samples are sorted in ascending order according  to the value of D. It can be found that when $D_l$ $<$ 25, i.e., 
sequence number less than 125, the three models are similar and all able to reproduce the pattern of true values. 
When 25 $<$ $D_l$ $<$ 50, i.e. sequence number between 125 to 175, only the ESRN model can provide a more precise 
estimation. $Model_22$ and $Model_23$ 
suffer from obvious underestimation.
When $D_l$ $>$ 50, i.e. sequence number over 175, all models underestimate while the prediction of ESRN model is the closest to the 
true value. This reveals a wider applicability and stability of the ESRN model. 

\begin{figure}[]
    \begin{centering}
    \includegraphics[width=0.55\linewidth]{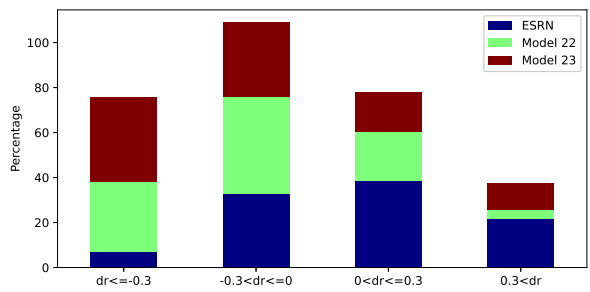}
    \caption{The DR distribution plot}
    \label{fig:DR}
    \end{centering}    
\end{figure}

The DR distribution for three models is shown in Fig. \ref{fig:DR}. For $Model_22$ and $Model_23$, 
DR is mainly located in $DR <= -0.3$ and $-0.3 <= DR <= 0$. 
This reveals that the prediction of $Model_22$ and $Model_23$ is more likely to be lower than the truth. 
And the main error is underestimation.   
On the contrary, the ESRN model tends to give slightly larger predictions, which can bring advantages in 
practical applications. That is because a 
larger estimation can offer more security redundancy, especially in the prediction of pollution transport.

\begin{figure}[]
    \begin{centering}
    \includegraphics[width=0.55\linewidth]{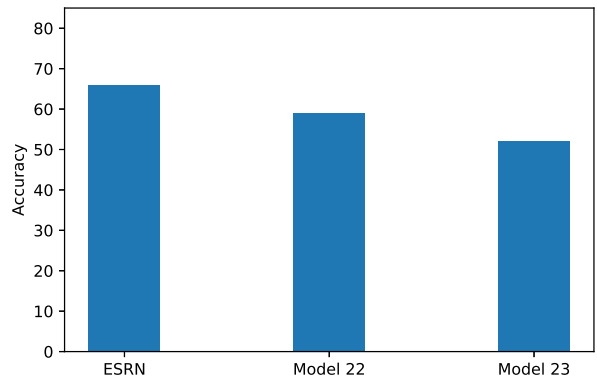}
    \caption{The accuracy plot}
    \label{fig:accuracy}
    \end{centering}    
\end{figure}

The accuracy plot reveals that the ESRN model has the highest accuracy (Fig. \ref{fig:accuracy}). Although 
the accuracy gap between the ESRN and $Model_22$ is small, 
ESRN model has better application potential considering the simplicity and lower demand for parameters.

\begin{figure}[]
    \begin{centering}
    \subfigure[Whole dataset]{
        \includegraphics[width=0.38\linewidth]{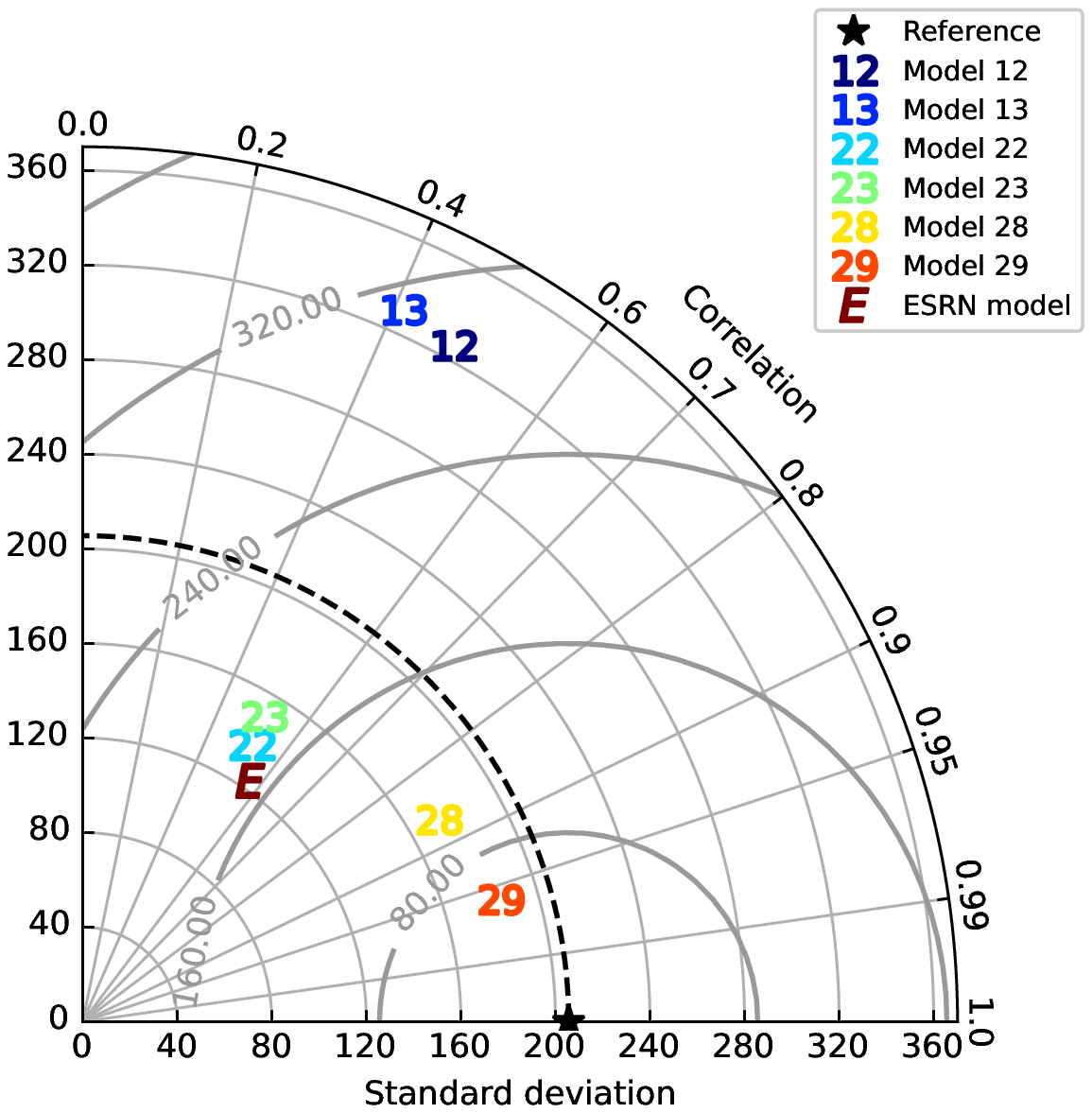}
    }

    \subfigure[Training dataset]{
        \includegraphics[width=0.38\linewidth]{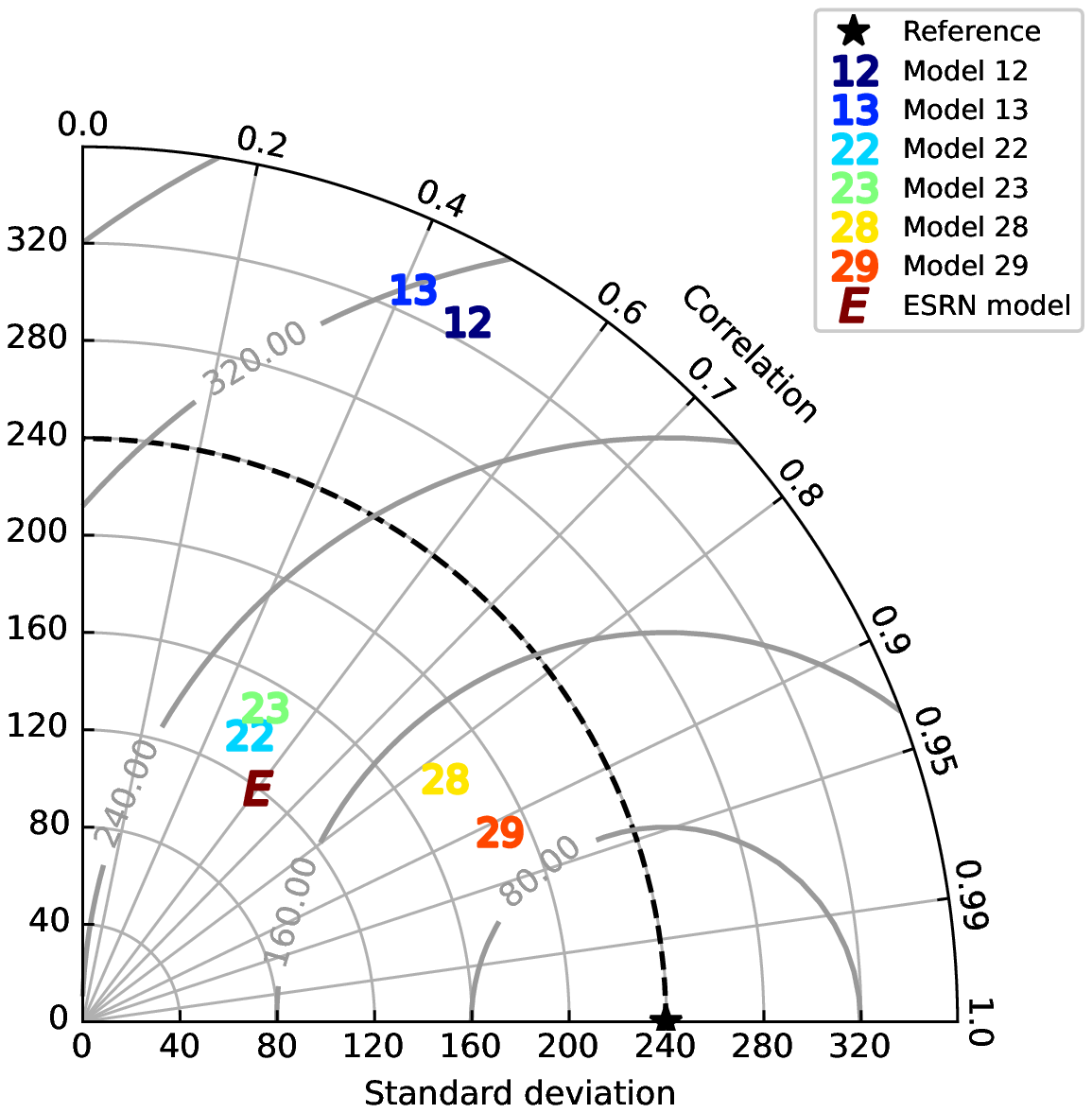}
    }

    \subfigure[Testing dataset]{
        \includegraphics[width=0.38\linewidth]{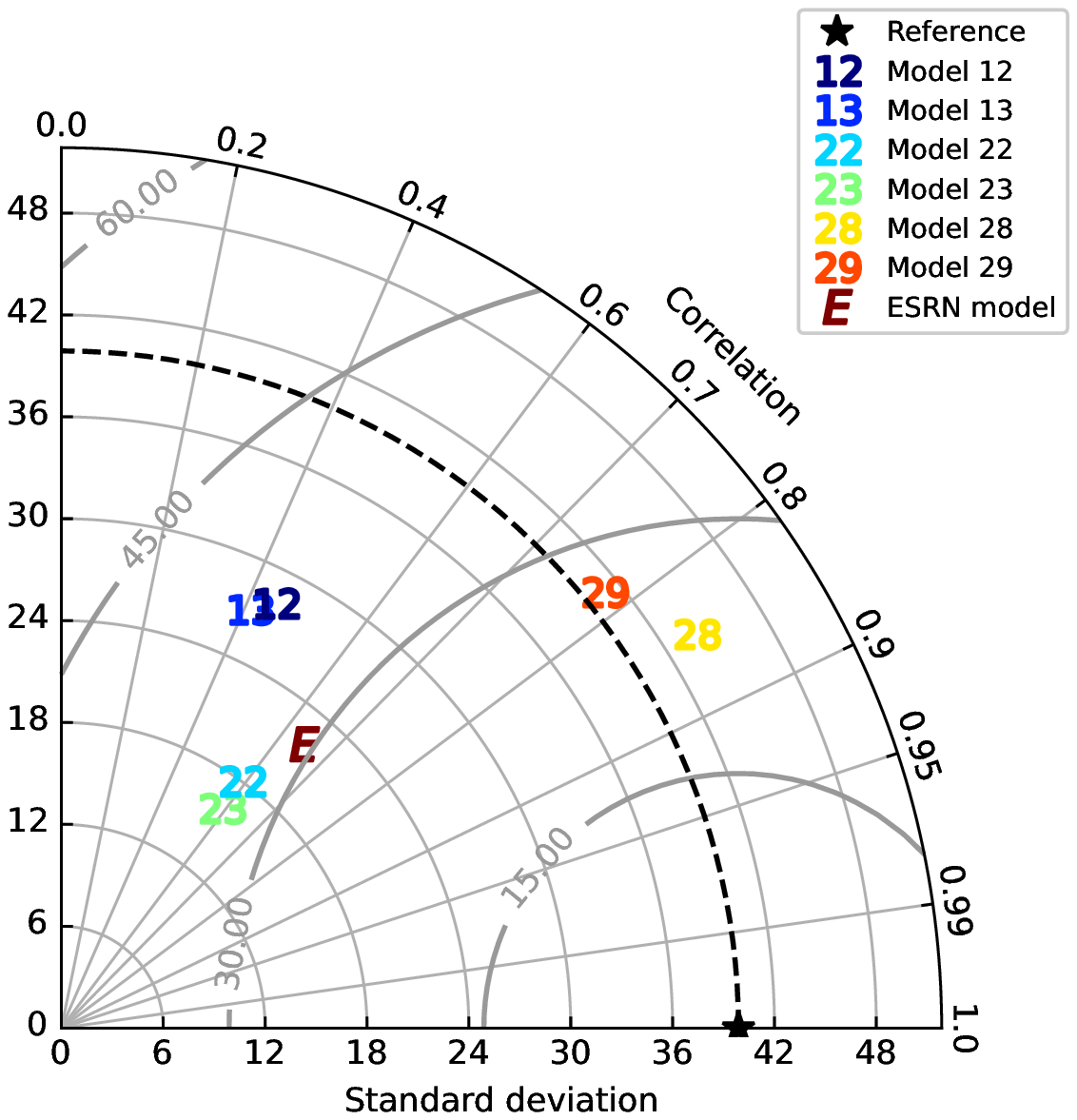}
    }

    \caption{The visual comparison of the ESRN model with other state-of-art models in each method-branches on whole, training and testing dataset }
    where Model 12 - Zeng and Huai / 2014; Model 13 - Disley et al.  / 2014; Model 22, 23 - Memarzadeh R., et al. / 2020; 
Model 28 - ElasticNet model; Model 29 - SVM model.
    \label{fig:SRcompare}
    \end{centering}    
\end{figure}

For more precise evaluation on different research methods, Taylor diagram of representative models from 
each method and ESRN model is plotted on whole, training and 
testing dataset. As shown 
in Fig. \ref{fig:SRcompare}, the distribution of those models is slightly different on distinct datasets, 
but the relative positions are in consistency.  
The overall performance of ESRN locates between model 3 and model 5, which means it has surpassed the state of art 
explicit ML-driven research with a much simpler form. 
It is clear that the ESRN model is now the superior symbolic model in the estimation of $D_l$. 
Although there is still a gap between ESRN and implicit models, its conciseness and explicability are enough to make up for this shortcoming. 
However, implicit ML-driven methods, model 28 and 29, still perform better due to their advantage in model complexity. 

In this equation, only $w$ and $U^*$ are involved in the final form. This is reasonable from two aspects: the performance advantage
and the variable correlation. The performance advantage is already illustrated above. As for the variable correlation, the SCC plot
has to be mentioned (Fig. \ref{fig:SCC}). It can be found that the relationship between $w$ and $d$(0.78) is strong. 
This indicates that only one 
of them is already able to express all the information on channel geometrics and both can represent as the characteristic length for the channel. 
Considering $w$ has a stronger connection to the D, 
it is proper to use only $w$. A similar relationship can be found between $U$ and $U^*$. But the situation is more complicated in 
stream properties. It is worth noting that $U^*$ has a much weaker relationship 
with $w$ and $d$ than $U$($w-U$=0.23, $w-U$=0.29 $>$ $w-U^*$=0.07, $w-U^*$=0.11), 
which reveals an independence between $U^*$ and $w$, $d$. Considering the strong connection between $U$ and $U^*$, the importance of $U^*$
is over $U$. Therefore, $w$ and $U^*$ in this dataset are the most important variables to prediction of $D_l$. 
A model of only $w$ and $U^*$ is reasonable. 
Moreover, the proposed equation is more consistent with the training data in parameter correlation than other general 
form formulas (Eq. \ref{eq:generalForm}). As shown in Fig. \ref{fig:SCC}, all parameters in the dataset are in positive correlation  
with $D_l$. But Due to the limitation of formula form, this relationship cannot be satisfied in most of other existing equations. 
Taking Model 7 (Liu et al./1977) for example, it is defined as: 

\begin{equation}
    D_{l}=0.18\frac{w^{2}U^{0.5}(U^{*})^{0.5}}{d}
    \label{E:model10}
\end{equation}

It is clear that $d$ is in negative correlation with $D_l$, which violates the parameter relationship in the dataset revealed by the SCC plot.
This violation is common in other alternatives, which makes the ESRN model become the most physically robust one of all LDC 
predictive models.  

Additionally, the proposed equation shares similarity with Taylor's work in 1954\cite{taylor1954the}. Taylor's 
formula is an analytical result 
for turbulent circular pipe flow when the proposed equation for natural streams. It is defined as:

\begin{equation}
D_l = 10.0au_*
\label{Eq:taylor1954}
\end{equation}
where $a$ - the radius of the pipe; $u_*$ - the shear velocity for circular pipe flow, equal to $(\frac{gaS}{2})^{0.5}$. 

The methods used to develop these two equations are worlds apart, yet their forms are astonishingly analogous. 
In the novel equation, $w$ and $U^*$ are used. The $w$ can analogy $a$ in theory as the characteristic length 
scale for the channel. 
Although the definition of $U^*$(Eq. \ref{Eq:defU})
is different from $u_*$, they are all used to describe the physical influence of velocity gradient,
signaling the strength of the shear stress and its influence on the mixture in the presence of flow 
eddies. 
\begin{equation}
    U^* = \sqrt{gRS_e}
    \label{Eq:defU}
\end{equation}
where $R$ = the hydraulic radius; $S_e$ = the slope of the energy grade line. 

This principle has been overlooked in many previous studies, probably due to their poor regression strategies 
and limited datasets. With proper treatment of a large dataset, a good division between testing and 
training sets, as well as the ESRN method, which explores exhaustively a broad range of functional forms for $D_l$, 
we have connected this principle to our model and recovered 
the turbulent mixture formula as the underlying mechanism in natural stream dispersion processes.

\section{Conclusions}\label{sec:conclusion}
In this present study, a simple ML-driven explicit model for the prediction of $D_l$ is proposed with a novel evolutionary symbolic 
regression network(ESRN). 
This equation is distilled from a strong data basis of 660 samples collected worldwide (at least increased by 20\% than other research), 
outliers cleaned through IQR and sets divided by SSMD. 
Basically, this model shares a similar parameter topology with other existing models (Eq. \ref{Eq: res}). With the search of the optimal 
parameter combination and the help of robust regression ability, the ESRN algorithm 
explores multiple symbolic combinations and obtains a globally optimal model with only two 
input arguments.  
Various performance metrics and visualization methods are used for further evaluation. 
As from the comparison, the ESRN model outperforms other existing models in most statistical indexes.

Although this model still suffers from accuracy loss for $D_l$ over 100, the statistical 
analysis on the cleaned, extensive dataset collected worldwide (Table \ref{T:statOfData}) reveals that 
the median and IQR of the overall dataset are 25.90 and 94.33. This indicates that the ESRN model already has the ability to make 
precise estimations for $D_l$ than existing alternatives in most application 
cases all over the world. The above evaluation also shows that the ESRN model tends 
to give slightly larger values of $D_l$ than the real ones, which
can bring security into its prediction. 

Furthermore, the analytical formula given 
by the ESRN is simple and represents the fundamental physics of turbulent mixing, a principle that has 
been overlooked in previous studies due to their poor regression strategies on smaller datasets.  

Additionally, the performance advantage also indicates that a larger dataset with better 
processing and suitable division between 
training and testing sets can offer a more accurate prediction less prone to overfitting. 

With advantages of simpler calculation, lower parameter-demand and higher accuracy, the ESRN model is very 
suitable for practical engineering problems. 
It can find its place in applications such as estimation of pollution spills, assessment of groundwater etc. 
The proposed ESRN algorithm performs outstandingly on this four-variable dataset. If a dataset with more input 
parameters can be provided, it might 
be able to find an even better variable combination and provide new prediction formulas with stronger performance. 
Further enhancement on the dataset is needed. 

\section*{Acknowledgement}\label{sec:Acknowledgement}
This work was financially supported by Westlake University and The 
Belt and Road Special Foundation of the State Key Laboratory of
Hydrology-Water Resources and Hydraulic Engineering (2019491511).

We also gratefully acknowledge Prof. Ling Li at Westlake University
for his useful comments on the manuscript.

\bibliographystyle{unsrt}  
\bibliography{references}

\end{document}